\documentclass[runningheads]{llncs}
\usepackage{graphicx}
\usepackage{pifont}
\newcommand{\cmark}{\ding{51}}%
\newcommand{\xmark}{\ding{55}}%
\begin{document}
\title{StandardSim: A Synthetic Dataset For Retail Environments}
\author{Cristina Mata\inst{1} \and
Nick Locascio \inst{2} \and
Mohammed Azeem Sheikh \inst{2} \and
Kenny Kihara \inst{2} \and
Dan Fischetti \inst{2}}

\authorrunning{C. Mata et al.}

\institute{Stony Brook University, Stony Brook NY 11790, USA \\
\email{cfmata@cs.stonybrook.edu} \\
Standard Cognition, 965 Mission St, San Francisco CA 94103, USA
\email{\{nick,mohammed,kenny,dan\}@standard.ai}\\
}

\maketitle              % typeset the header of the contribution
\begin{abstract}
Autonomous checkout systems rely on visual and sensory inputs to carry out fine-grained scene understanding in retail environments. Retail environments present unique challenges compared to typical indoor scenes owing to the vast number of densely packed, unique yet similar objects. The problem becomes even more difficult when only RGB input is available, especially for data-hungry tasks such as instance segmentation. To address the lack of datasets for retail, we present StandardSim, a large-scale photorealistic synthetic dataset featuring annotations for semantic segmentation, instance segmentation, depth estimation, and object detection. Our dataset provides multiple views per scene, enabling multi-view representation learning. Further, we introduce a novel task central to autonomous checkout called change detection, requiring pixel-level classification of takes, puts and shifts in objects over time. We benchmark widely-used models for segmentation and depth estimation on our dataset, show that our test set constitutes a difficult benchmark compared to current smaller-scale datasets and that our training set provides models with crucial information for autonomous checkout tasks.

\end{abstract}

\section{Introduction}
Autonomous checkout is a fast-spreading technology poised to change the way customers shop in brick and mortar stores. It often relies on cameras and other sensors to build an understanding of a retail environment and make a final decision about what a shopper purchases. Computer vision plays a crucial role in understanding this data, especially in systems where cameras are the only sensors available. While vision-only autonomous checkout is relatively new, progress in the domain has not led to any benchmarks or new tasks. We hypothesize that understanding of retail environments not only requires data that is tailored to this domain, but also a new computer vision task that identifies changes in retail scenes over time. Thus, in this paper, we describe a new dataset StandardSim, as well as a new task that identifies changes in retail scenes over time.
% \par Autonomous checkout scenes present cluttered environments with many small objects that are difficult to identify from distant viewpoints. Object detection, semantic and instance segmentation, and depth estimation are tasks that may be useful in identifying small objects in retail environments since they produce pixel-level understanding of objects in a scene. Models trained for these tasks require vast amounts of labeled data to produce accurate results, yet there is a lack of datasets tailored for retail environments that include annotations for these tasks. 
\par While object detection in retail environments has been broached by \cite{goldman2019dense}, which introduced the dense object detection task and dataset SKU-110K in the retail setting, they do not provide semantic annotations of objects beyond bounding boxes. Other large-scale datasets have been created synthetically for indoor environments, but they do not transfer well to retail because objects are not as diverse or densely packed. Additionally, these datasets do not provide diverse viewpoints, often showing scenes from the perspective of a person navigating the scene, as opposed to from the ceiling or a corner. Thus, when models trained on these datasets are applied to real retail environments they have not learned information needed to identify small objects from ceiling camera perspectives.
% \par Additionally, to build a comprehensive understanding of a shopper's actions in a retail environment, the standard computer vision tasks of object detection, segmentation and so on do not deliver information tailored to downstream decisions about a shopper's cart. These decisions are oriented around changes to the environment, while current tasks focus on understanding a single frame and ignore the temporal dimension.
\par The change detection task is meant to simulate a shopper's actions in a retail environment by providing a model with a pair of images of a scene, before and after a series of interactions with objects in the scene. After a shopper interacts with objects in a retail environment, objects may be taken, added or shifted around the scene. Each image pair in our dataset displays what a random interaction might look like and is annotated similarly to segmentation, where each pixel belongs to a take, put, shift or no change. Objects tend to be small and changes are sparse. Due to the task's similarity to segmentation, we adapt a popular state of the art segmentation model, Deeplabv3 \cite{deeplabv3plus2018}, to set a benchmark for this task, and show that due to the sparse changes and small size of objects, StandardSim is a very difficult benchmark.
\par With these problems in mind, we introduce StandardSim, a large-scale synthetic dataset made from highly accurate store models and featuring annotations for depth estimation, object detection, instance segmentation and a novel task which we call change detection. This dataset consists of over 25,000 images from 2,134 unique scenes. Each scene includes views from multiple cameras, enabling multi-view reconstruction and the generation of shape estimation annotations. Compared to previous datasets, StandardSim provides annotations for more tasks and fills a void in the retail environment domain. 
\par In addition to change detection we focus on monocular depth estimation because depth provides important cues about the movement of objects over time. We benchmark the state of the art monocular depth estimation model Dense Prediction Transformer \cite{Ranftl2021}, based on MiDaS \cite{Ranftl2020}, on StandardSim and compare its performance on other datasets. We find that it has a much higher error on our dataset, suggesting that StandardSim constitutes a difficult new benchmark for monocular depth estimation. StandardSim will be made accessible via URL post-publication.
% \par To summarize, we present the following contributions:
% \begin{itemize}
%     \item We introduce StandardSim, a large-scale synthetic dataset with photorealistic store layouts and objects that includes annotations for object detection, instance and semantic segmentation, surface normal segmentation, and monocular depth estimation.
%     \item We introduce the change detection task for retail, where images before and after a shelf interaction must be labeled at a pixel level to denote what objects are taken, put and shifted.
%     \item We present benchmarks for change detection and monocular depth estimation using widely used state of the art models and show that our dataset constitutes a difficult benchmark for both tasks. Finally we discuss the implications of our dataset for the domain of retail environments, which has been largely ignored in the literature up to now.
% \end{itemize}
\section{Related Work}
\subsection{Datasets for Retail Environments}
So far work on computer vision for retail environments has focused on object detection in densely packed scenes \cite{goldman2019dense}. In these environments, traditional object detection solutions fail because the boxes are either inaccurate or overlap to a degree that makes them less useful. In order to address this, the authors of \cite{goldman2019dense} released a new dataset, SKU-110K, that includes densely packed views of shelves that are annotated with bounding boxes around the objects (typically called stock-keeping units (SKUs) in the retail environment). For this dataset, the primary limitations are that the data consists of front facing views of the objects only and the objects are only annotated with bounding boxes. Our synthetic dataset does much to address some of these shortcomings, including providing a variety of shelf styles (such as hanging items), pixel level masks for each object in the image, and different, skewed camera perspectives for each object.
\subsection{Monocular Depth Estimation}
Another shortcoming of retail datasets is the lack of depth information, which is key to both detect the layout of stores and to detect potential changes. Our dataset includes depth maps for each image, as well as multiple camera views to allow for reconstruction through multiple view points. Current state of the art in monocular depth estimation includes zero shot approaches \cite{Ranftl2020} trained on multiple datasets to adapt well to various environments. Other approaches combine semantic segmentation and instance information with RGB images to obtain monocular depth maps \cite{Wang_2020_CVPR} \cite{Saeedan_2021_WACV} and using 360 views combined with edge and corner information to obtain good indoor depth maps \cite{Jin_2020_CVPR}. Combining instance segmentation, semantic segmentation, and monocular depth estimation has also been explored \cite{Goel_2021_WACV}. Finally, domain adaptation from synth-to-real is becoming increasingly important \cite{Zhao_2020_CVPR}, since many real world applications require accurate depth maps with no easy way to measure ground truth.
\subsection{Change Detection}
In addition to the work on retail environments, we also look at a new task, that of visual change detection. Visual change detection has been explored for outdoor urban scenes \cite{Varghese_2018_ECCV_Workshops}, which is similar to the task definition we provide, except that our scenes are indoors with labels tailored to retail settings. Other modalities for change detection, such as a purely classification based approach \cite{liu_urban_2002}, exist in the urban change detection area. However, not many datasets exist. In addition, a deep literature exists for change detection in remote sensing applications \cite{wenzhong_shi}, along with many open source datasets.
\par In addition to outdoor and remote sensing tasks, the new work ChangeSim \cite{park2021changesim} highlights change detection in an indoor setting. Several differences exist between this work and our task. Our cameras are fixed for a given set of changes, and our images are paired, with a well-defined before and after image. Our dataset also consists of retail environments rather than industrial environments. Finally we have multiple camera views per change, allowing for the potential implementation of multi-camera change detection. However, we do share many similarities. Both tasks involve change detection using semantic masks in challenging environments, in which background characteristics such as lighting are in flux.
\subsection{Synthetic Indoor Datasets}
Besides ChangeSim \cite{park2021changesim}, other photorealistic synthetic datasets exist. Two prominent datasets are Hypersim \cite{roberts} and SceneNet \cite{McCormac}. Both focus heavily on several important aspects of scene understanding, with a focus on semantic segmentation, object detection, and depth estimation. These datasets are often composed of highly detailed indoor environments, but do not include retail environments. They have several pieces in common with our dataset, including varied lighting and different textures available for each scene, as well as detailed semantic labels for each object in the image.
\section{StandardSim}
\subsection{Overview}
%%%%%%%%%%%%%%%%%%%%%%%%%%%%%%%%%%%%%%%%%%%%%%%
% Figure showing a StandardSim sample
%%%%%%%%%%%%%%%%%%%%%%%%%%%%%%%%%%%%%%%%%%%%%%%
\begin{figure}[!t]
\caption{From left to right, an image sample with corresponding random matting on the shelves, depth annotation, and surface normal annotation.}
\label{fig:annotation_sample}
\tabcolsep 0.03cm
\noindent\makebox[\textwidth]{%
\begin{tabular}{cccc}
  \includegraphics[trim = 0mm 0mm 0mm 0mm,clip,height=1.5cm]{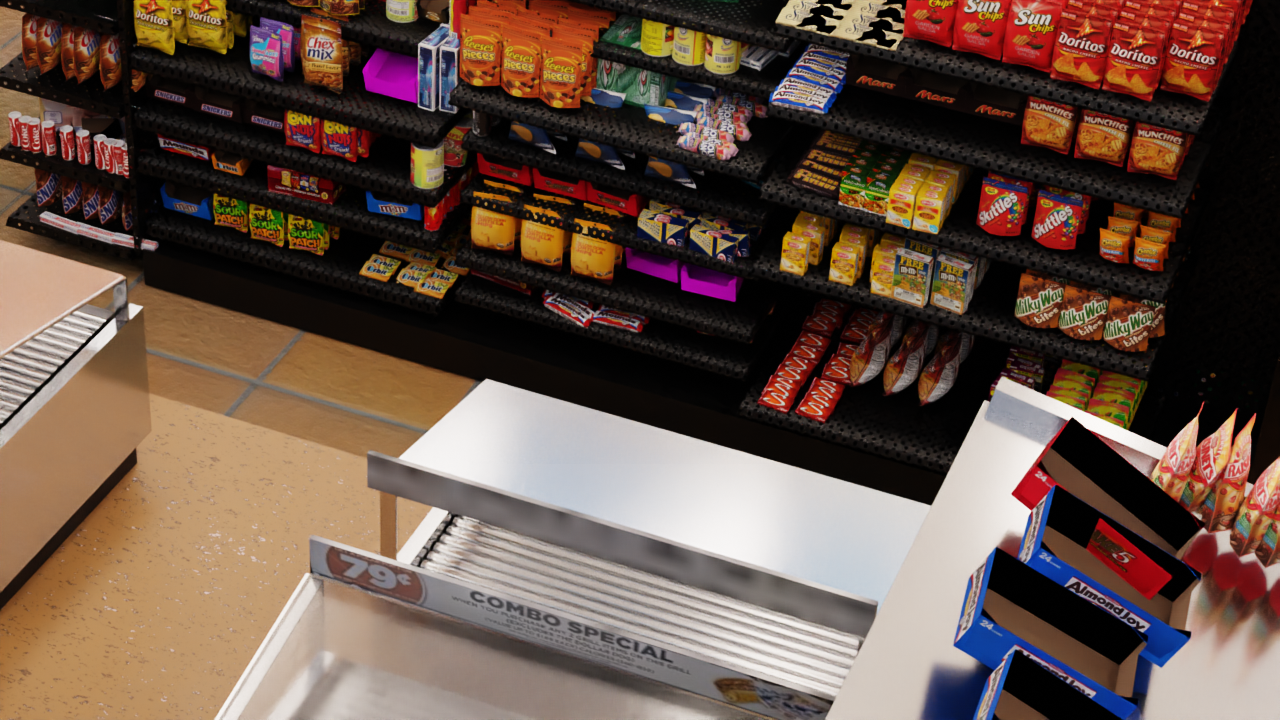} &
  \includegraphics[trim = 0mm 0mm 0mm 0mm, clip, height=1.5cm]{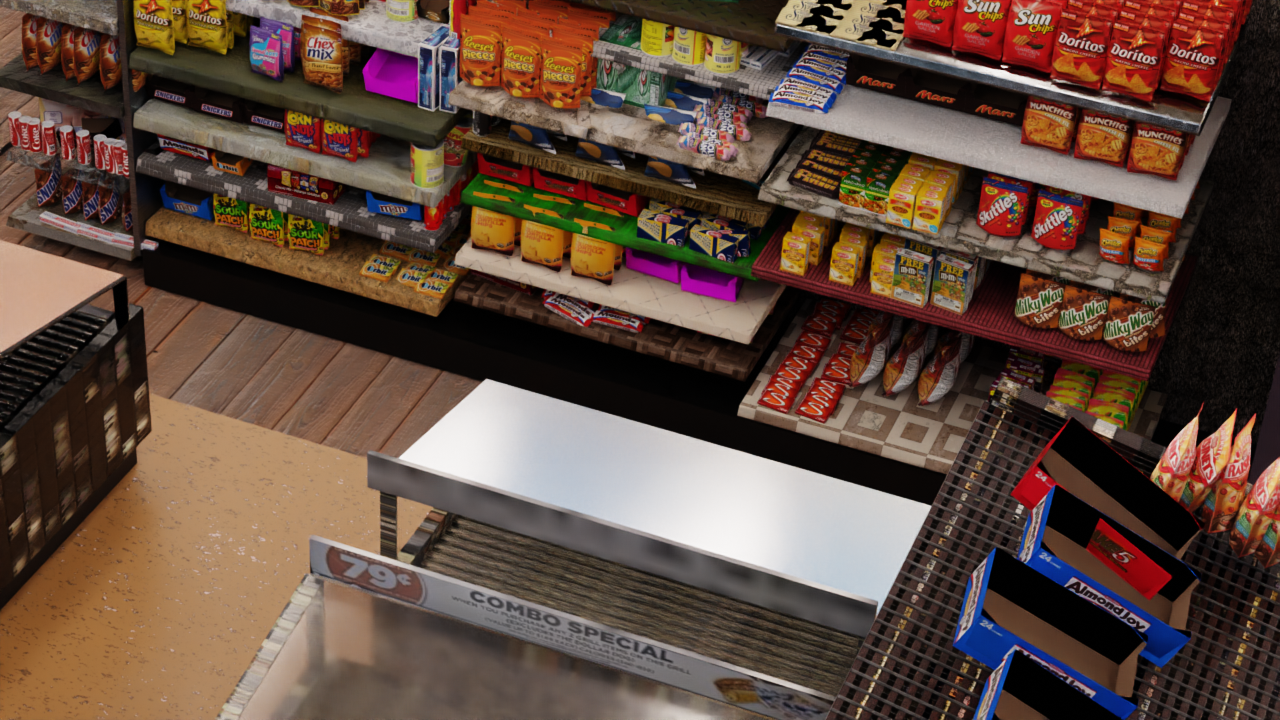} &
  \includegraphics[trim = 0mm 0mm 0mm 0mm,clip,height=1.5cm]{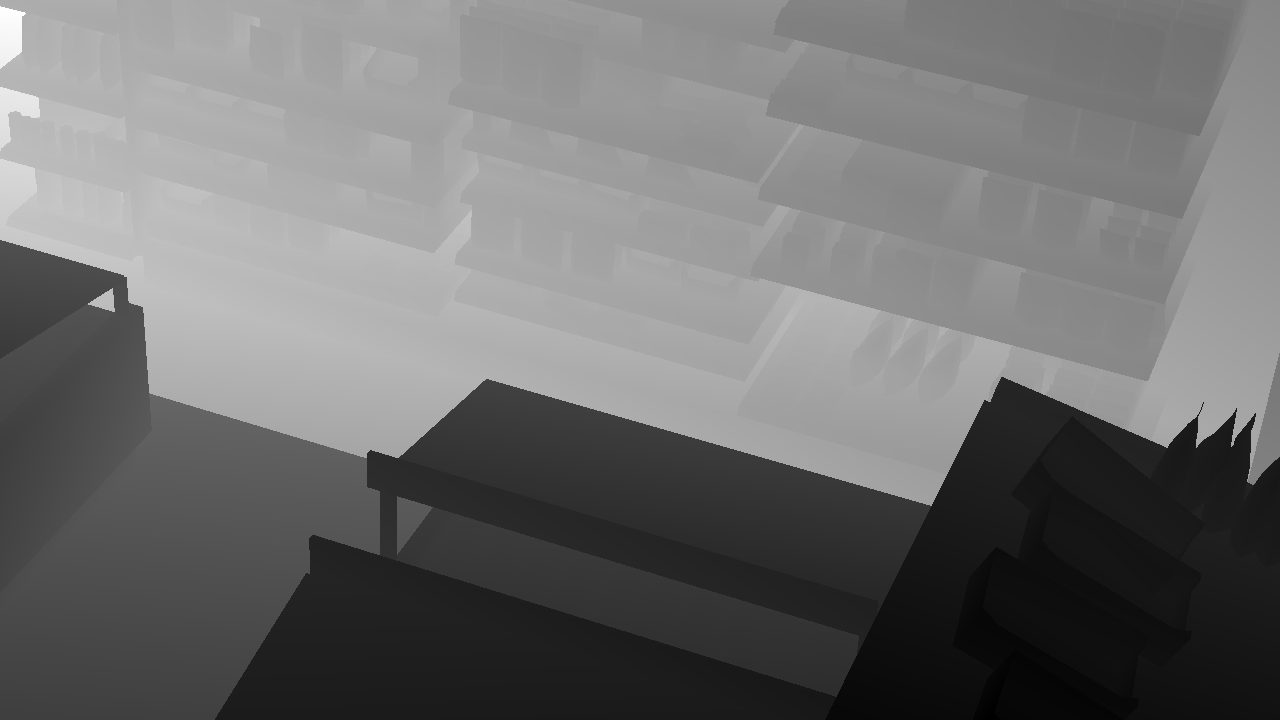} &
  \includegraphics[trim = 0mm 0mm 0mm 0mm, clip, height=1.5cm]{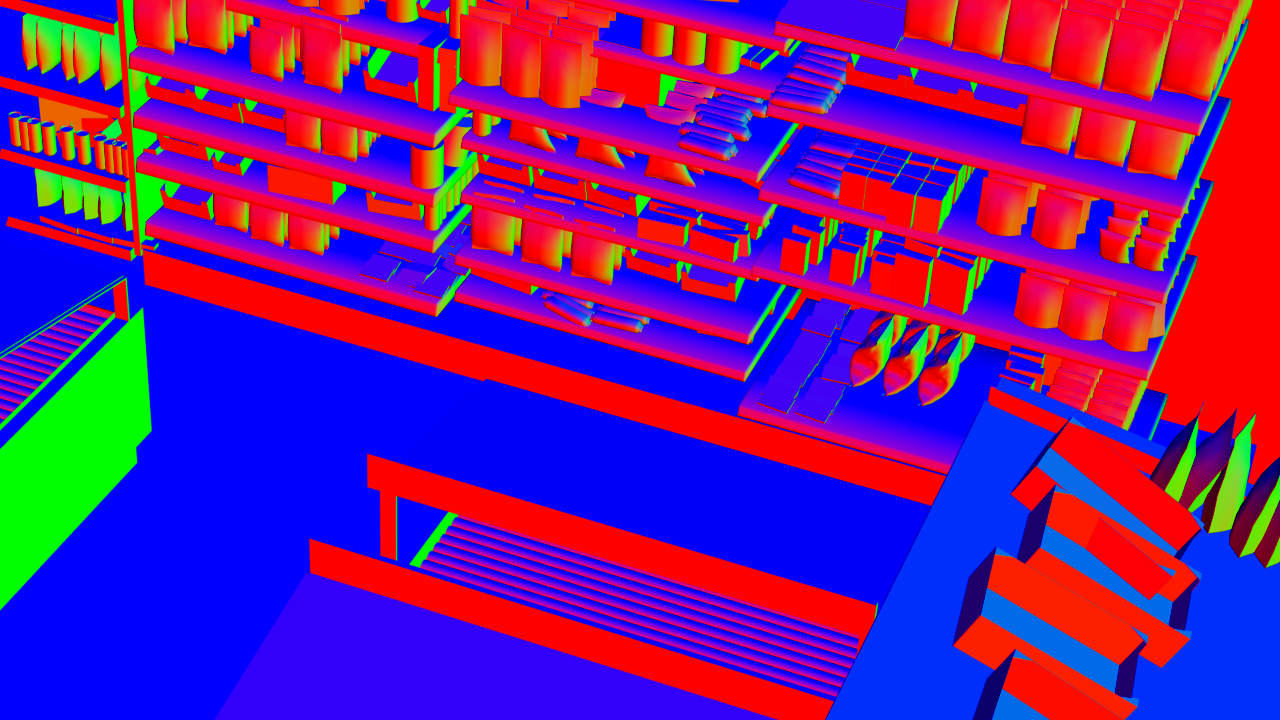} \\
\end{tabular}}
\end{figure}
\par
\begin{figure}[!t]
\caption{Store layouts used in the dataset. Best viewed in color and zoomed in.}
\label{fig:store_layouts}
\makebox[\textwidth]{
\begin{tabular}{ccccc}
  \includegraphics[trim = 0mm 0mm 0mm 0mm,clip,height=1.2cm]{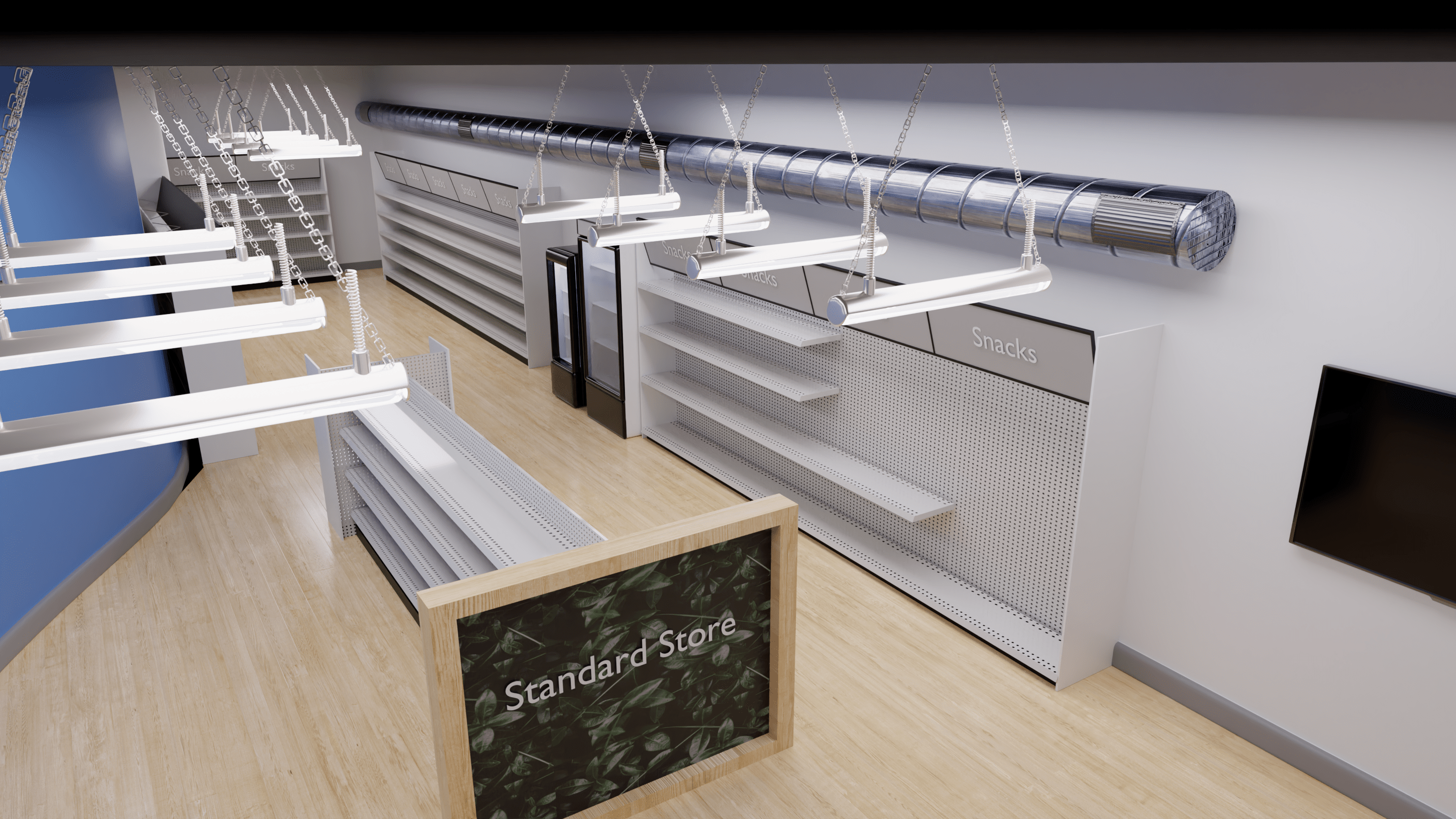} &
  \includegraphics[trim = 0mm 0mm 0mm 0mm,clip,height=1.2cm]{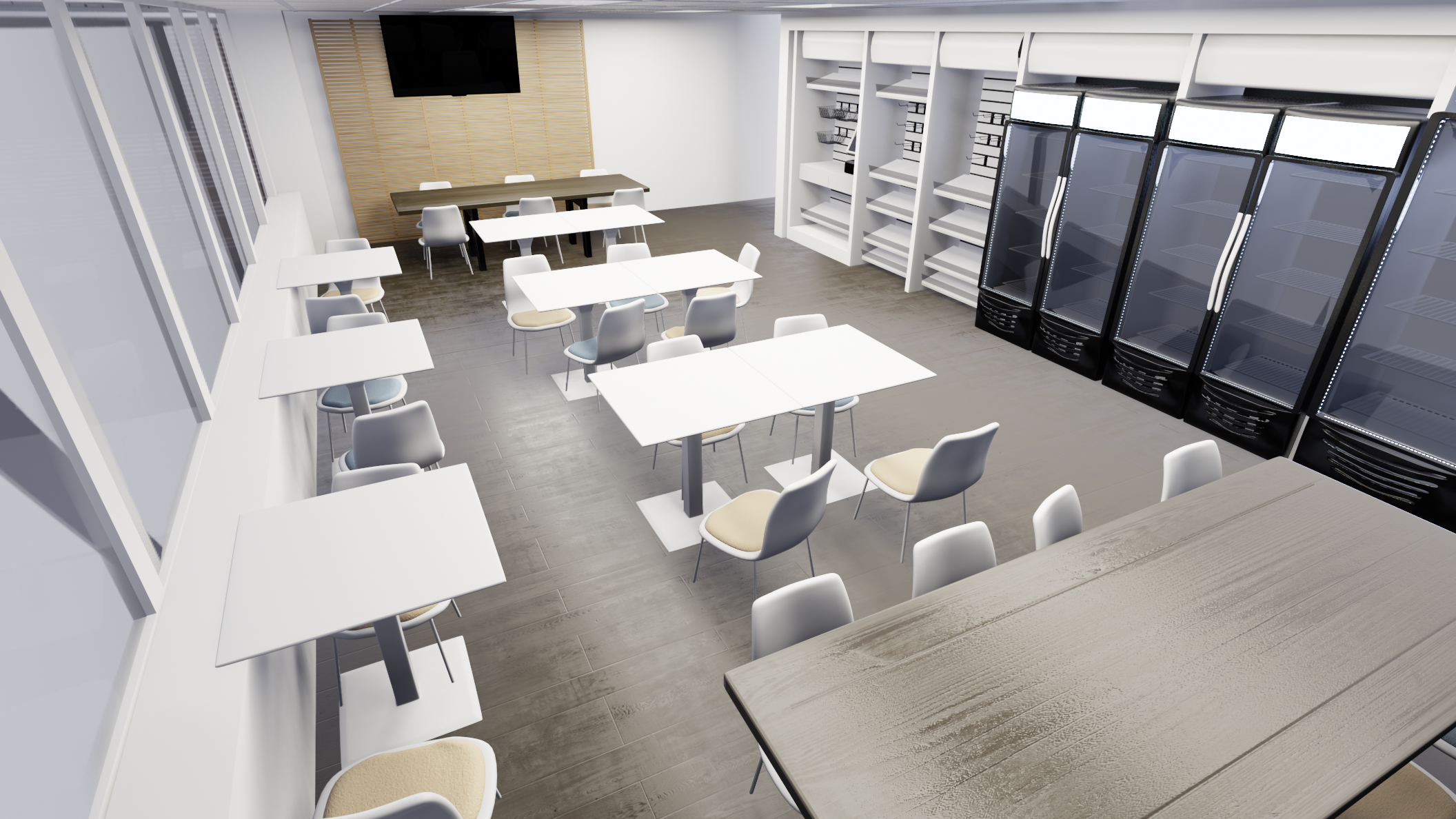} &
  \includegraphics[trim = 0mm 0mm 0mm 0mm, clip, height=1.2cm]{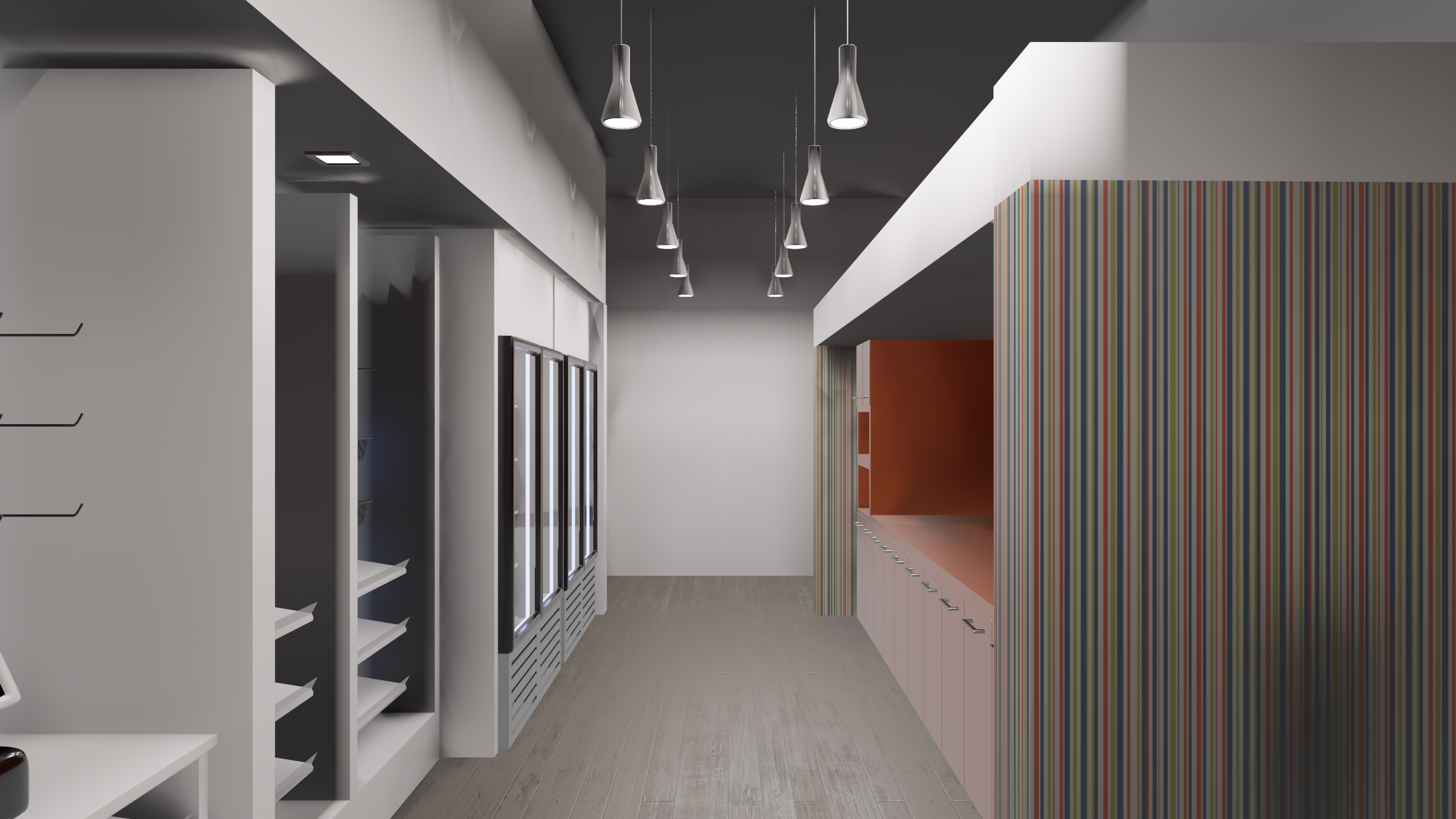} &
  \includegraphics[trim = 0mm 0mm 0mm 0mm,clip,height=1.2cm]{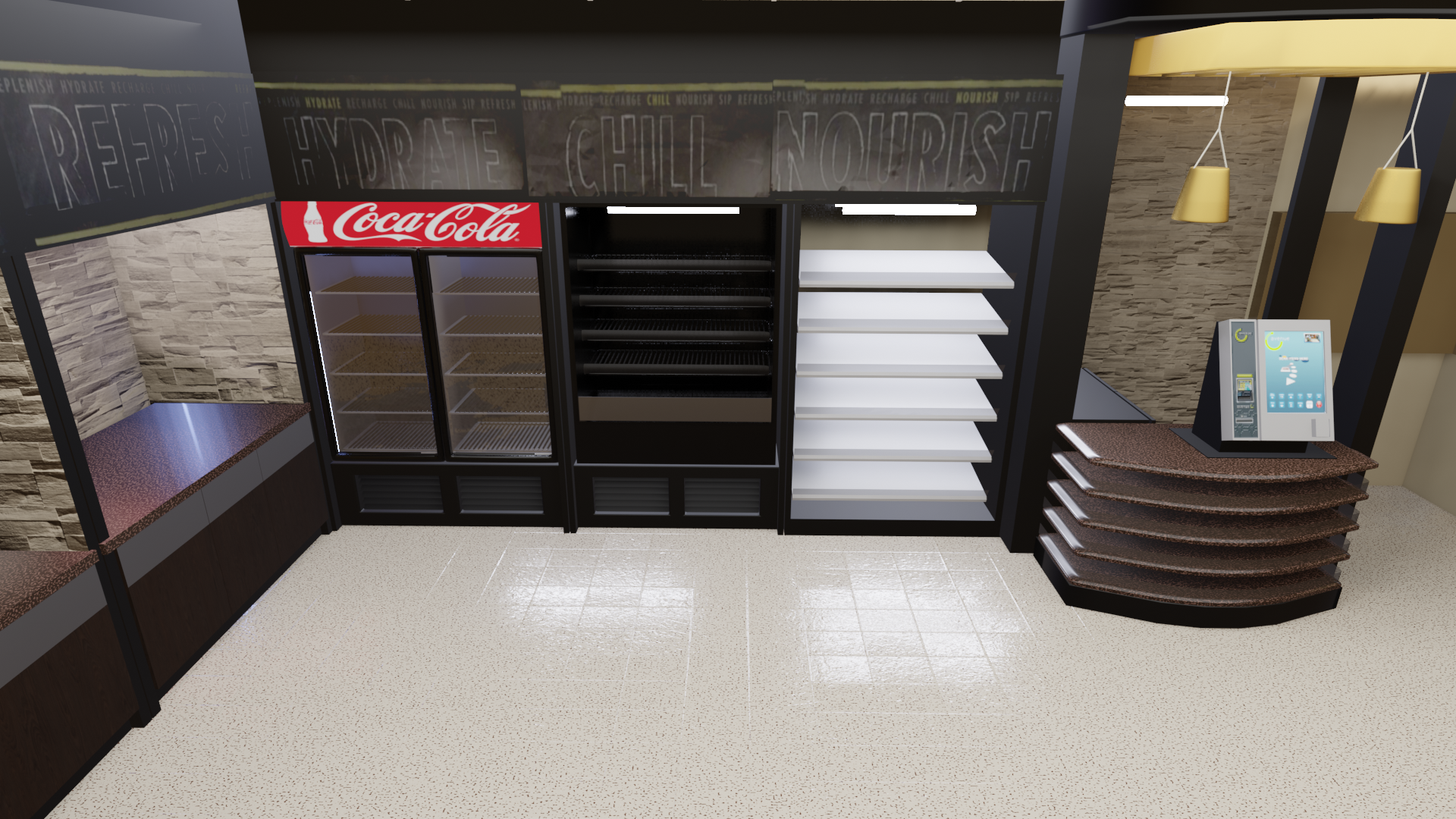} &
  \includegraphics[trim = 0mm 0mm 0mm 0mm,clip,height=1.2cm]{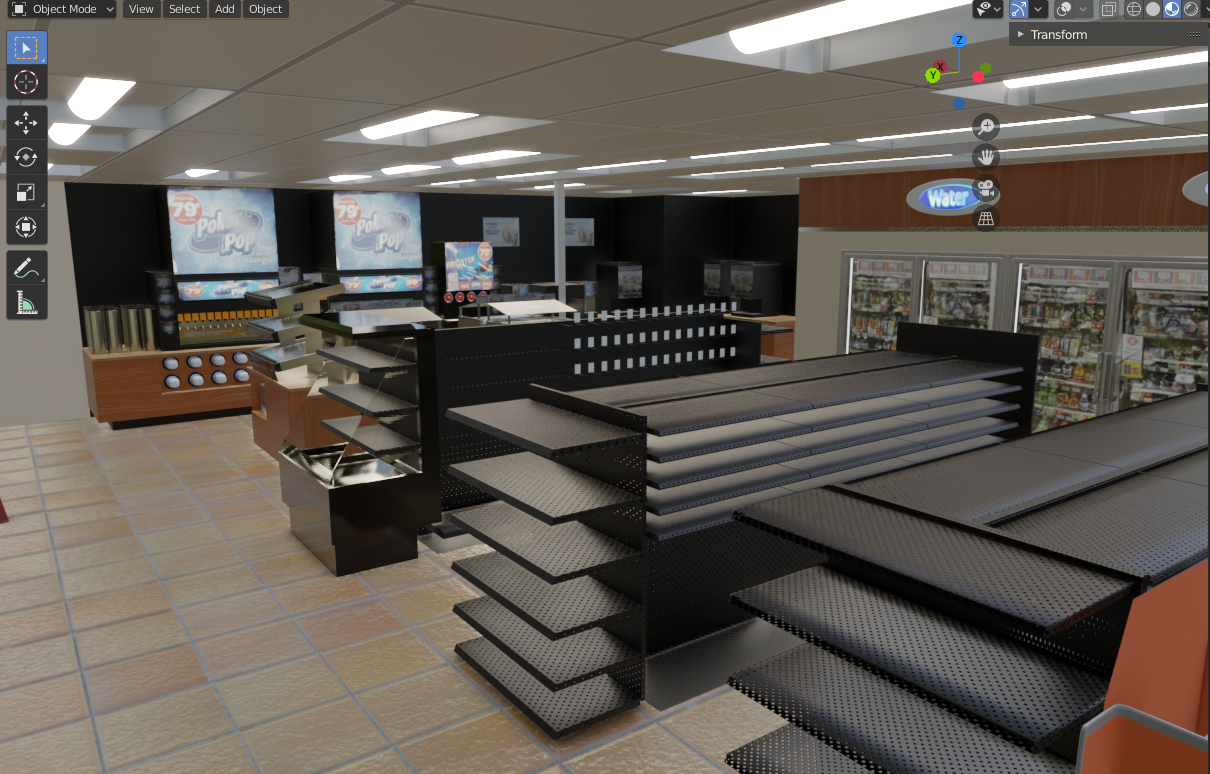} 
\end{tabular}
}
\end{figure}

 Our dataset consists of 2,134 unique scenes of retail environments based on three real store layouts. Each scene consists of a set of actions in which several items may be taken, put or shifted. There are also scenes in which no changes occurred, which reflects real-world retail environments. Our dataset is focused on the change detection task which aims to discern what changes occur in a scene at a pixel level, so we structure a sample as a pair of before and after images. In total our dataset contains 6,359 such samples. We also add random matting to the shelves, resulting in two distinct mattes for each sample. Thus we have 4 images per sample from 2 different mattes, leading to 25,436 images in total. Figure \ref{fig:store_layouts} shows views of the empty stores in good lighting so that all details are viewable. It features diverse shelving and lighting configurations, as well as reflective surfaces from glass. 
\par We chose a 70/15/15 training, validation and testing split. We randomly assigned scenes while keeping the distribution of samples across stores as even as possible, that is, we have 1/3 of the samples from each split come from each store. Additionally, there is an average of 3 camera views per scene to provide data for multi-view tasks.

% One can find a comprehensive breakdown of the number of scenes and samples from each split in Table \ref{tab:dataset_breakdown_splits}.

\par In Table \ref{tab:dataset_comparison_breakdown} we show that compared to other datasets for retail and change detection, our dataset features more images and annotations for more tasks. Other large-scale synthetic datasets may provide more images but they do not provide change detection annotations and do not structure their samples for change detection. Our dataset provides scenes more suitable for autonomous checkout.
%%%%%%%%%%%%%%%%%%%%%%%%%%%%%%%%%%%%%%%%%%%%%%%
% Table with dataset splits
%%%%%%%%%%%%%%%%%%%%%%%%%%%%%%%%%%%%%%%%%%%%%%%
% \begin{table}[!t]
% \centering
% \setlength\tabcolsep{4pt}
% \caption{Breakdown of samples by split}
% \label{tab:dataset_breakdown_splits}
% \begin{tabular}{|l|l|l|l|l|}
% \hline
%  & Train & Val & Test & Total \\
% \hline
% \# Scenes & 1493  & 320 & 321 & 2134 \\
% \# Samples & 4455 & 949 & 955 & 6359 \\
% \hline
% \end{tabular}
% \end{table}
%%%%%%%%%%%%%%%%%%%%%%%%%%%%%%%%%%%%%%%%%%%%%%%
% Table with StandardSim comparison
%%%%%%%%%%%%%%%%%%%%%%%%%%%%%%%%%%%%%%%%%%%%%%%
\begin{table}[!t]
\caption{Comparison of StandardSim to other datasets. The columns refer to the type of annotation provided: from left to right these are Depth, Object Detection, Instance Segmentation and Change Detection annotations. The Domain refers to the type of locations displayed in the dataset.}
\label{tab:dataset_comparison_breakdown}
\centering
\begin{tabular}{|l|l|l|l|l|l|l|}
\hline
 & Depth & Object Det. & Instance & Change Det. & Domain & \# Images \\
\hline
ChangeSim & \cmark & \xmark & \xmark & \cmark & Industrial & 130,000 \\
SKU-110K & \xmark & \cmark & \xmark & \xmark & Retail & 11,762 \\
HyperSim & \cmark & \cmark & \cmark & \xmark & Indoor & 74,619 \\
SceneNet & \cmark & \cmark & \cmark & \xmark & Indoor & 5M \\
StandardSim & \cmark & \cmark & \cmark & \cmark & Retail & 25,436 \\
\hline
\end{tabular}
\end{table}
%%%%%%%%%%%%%%%%%%%%%%%%%%%%%%%%%%%%%%%%%%%%%%%
% Figure showing store layouts
%%%%%%%%%%%%%%%%%%%%%%%%%%%%%%%%%%%%%%%%%%%%%%%
\subsection{Dataset Creation}
We utilize Blender \cite{Blender} for the core of our data generation pipeline. Our data generation pipeline utilizes Blender's python interface to dynamically modify the store, the products, and the camera placements. We utilize Blender's cycles render engine to produce photo-realistic RGB, RGB with randomized textures, RGB with blank textures, z-depth, segmentation masks, and surface normals. Each image render takes approximately 20 seconds utilizing 2 GPUs (Tesla V100 16GB). With the publication of this paper the generation and rendering pipelines will be made available and open-source.
\par Each store model is a to-scale replica of a real retail store. To create these replica assets, we perform a 3d scan of the store using a lidar-based Matterport \cite{noauthor_pro2_nodate} device, and utilize an in-house asset creator to model the store in Blender with the 3d scan as a guide. This ensures highly accurate dimensions and structure of the 3d model, while also producing high quality textures and meshes required for photo-realistic rendering from many viewpoints. We model the store and indicate which shelves meshes are shelves, but leave all shelves empty.
\par For our SKU models and textures we purchased a variety of assets from the online marketplace Turbosquid \cite{noauthor_3d_nodate}. We build a collection we call SkuBank of 456 assets with high quality meshes and textures as .obj, .mtl, and .png files. Due to the licensing restrictions of these acquired assets, they are not part of our release. However, our data generation pipeline is flexible to any SkuBank built in this way, so substituting other objects can be done in a plug-and-play fashion. We place the objects from SKUBank on the store shelves in an automated, random fashion that aims to replicate the densely packed distribution seen in retail environments.
\par Once a store's product placement is complete, we focus on generating random, but valid extrinsic and intrinsic camera parameters. We supply these camera parameters in standard opencv matrix format \cite{opencv_library}. Intrinsic parameters are fixed other than the field-of-view of the camera which we sample from the range [50, 70]. Extrinsic camera placement in world coordinates follows a simple algorithm, CamPlace. The camera height (z coordinate) is placed on the ceiling with some random height adjustments. The camera (x,y) global coordinates are determined by sampling from a 5x5 meter plane around a random, front-facing sku. To filter out degenerate camera positions, we utilize Blender's built-in raytracing to ensure that the object in question is visible from the camera. This ensures that the camera is placed within the store, in free space, and avoids issues around the camera spawning inside of objects which may block its view. 

% The algorithm in further detail can be found in the figure below.

% \begin{algorithm}[H]
%  \KwData{SkuBank, ceiling, camera}
%  \KwResult{CamExtrinsics}
%  \While{True}{
%  camera.z = ceiling.z - random(0, 2) \\ % is this random uniform?
%  front\_sku = SkuBank.select\_random\_front\_sku() \\
%  camera.x = front\_sku.x + random.uniform(-5, 5) \\
%  camera.y = front\_sku.y + random.uniform(-5, 5) \\
%  camera.roll = random.uniform(-pi, pi) \\
%  off\_center\_sku\_pos = sku.position + random.uniform\_xyz(-1, 1)
%  camera.point\_at(off\_center\_sku\_pos) \\
%  \If{camera.ray\_trace\_visible(sku)} {
%   return camera.extrinsics
%   }
%  }
%  \caption{Psuedocode for the CamPlace algorithm}
% \end{algorithm}
\par In order to generate data for our change detection task, we must perturb the items in a realistic and principled way. Our change creation algorithm, SkuRemove, selects a random 2-10\% of product sections, and then selects a random (1-6) number of products items to remove from that section. When a customer takes a product from a section, nearby products often shift around it due to their proximity. We model this directly by randomly shifting nearby SKUs with a higher probability than the random shifting applied to all SKUs.
%%%%%%%%%%%%%%%%%%%%%%%%%%%%%%%%%%%%%%%%%%%%%%%
% Change Detection Task
%%%%%%%%%%%%%%%%%%%%%%%%%%%%%%%%%%%%%%%%%%%%%%%
\section{Change Detection Task}
\par We formulate the change detection task in the context of autonomous checkout, where it is crucial to know how objects have moved or shifted over time to put together a shopper's receipt. In our problem formulation, our input is a pair of images, $(X_{t_{1}}, X_{t_{2}})$, where $t_{1} < t_{2}$ and $t_{1}, t_{2}$ are distinct time stamps. Our goal is to output a pixel-wise labeling $Y = \{y\}_{1}^{N}$ where $N$ is the number of pixels in the frame and $y \in \{$ no change, take, put, shift $\}$. These classes represent the basic actions we are interested in, and in other applications such as industrial ones have been extended to include rotation \cite{park2021changesim}, a more specific semantic label than shift.
\par We follow the COCO-style annotation formatting \cite{lin2014microsoft} to provide annotations enabling change detection, semantic and instance segmentation, object detection and depth estimation. We separate image indexing from annotations for ease of use. 
%Below we show how a change detection sample is structured, representing meta-information with an ellipsis.
%\begin{lstlisting}[language=json,numbers=none]
%"images": [
%    {
%        ...
%        "randommat1": "'before' randommats image",
%        "randommat2": "'after' randommats image",
%        "depth1": "'before' depth image",
%        "depth2"" "'after' depth image",
%        "image1": "'before' image",
%        "image2": "'after' image"
%    }
%],
%"annotations": [
%    {
%    ...
%    "action": "action",
%    "segmentation": [[list of points for Polygon]]
%    "scene": "scene_id_cam-i",
%    "bbox": [x, y, w, h]
%    }
%]
%\end{lstlisting}
% Optional Figure: Show how two images are input to a model with an architecture for segmentation and the output shows all the different classes.
%%%%%%%%%%%%%%%%%%%%%%%%%%%%%%%%%%%%%%%%%%%%%%%
% Experiments
%%%%%%%%%%%%%%%%%%%%%%%%%%%%%%%%%%%%%%%%%%%%%%%
\section{Experiments}
\subsection{Change Detection}
Change detection requires a pixel-level understanding of where changes occur in a scene, leading us to choose a popular semantic segmentation model, Deeplabv3 \cite{deeplabv3plus2018}, to benchmark on our dataset. We chose the Resnet50 \cite{He2015} backbone as our architecture and found this provides a good trade-off of accuracy versus computational efficiency. We load COCO \cite{lin2014microsoft} pretrained weights into the encoder and finetune the entire network on our training set until convergence. We use a batch size of 16, the SGD optimizer with an initial learning rate of 0.0004 and cosine decay. We apply random crops and resizing to height 360 and width 512. All our experiments are conducted using PyTorch \cite{NEURIPS2019_9015}.
% This constitutes a 2x downsample from the original size, which decreases memory usage without sacrificing accuracy.
\par We include several data augmentations specific to the change detection task. First we randomly modify change detection pairs by flipping the change order, that is, we switch the before and after images and keep the same label. Second, we randomly apply left-right flipping to both images and labels. We also randomly add noise to pixel values to simulate real camera images. In addition, the images include changes in lighting that might occur in real camera feeds.
\par During training we measure the mean Jaccard Index or Intersection over Union (IOU) on the validation set and stop training when this metric plateaus at 1950 epochs. In our initial experiments, the model experienced mode collapse predicting all background, but we found that adding loss weights alleviates this issue. Our final model achieves 36.15\% IOU on the validation set and 36.04\% IOU on our test set, indicating that change detection is a very difficult task similar to fine parts segmentation. A breakdown of IOU by class can be found in Table \ref{tab:change_det_results_breakdown}, which shows that the model performs best on the put class and struggles most on the shift class. These results also suggest that our test set is unbiased relative to our validation set due to the similarity in results between the two. Figure \ref{fig:results_change_detection} shows how the model performs qualitatively on our dataset, showing good results on even small objects. 

% However the model still struggles with dim lighting and object boundaries as can be see in in Figure \ref{fig:change_detection_failure}.

\par Additionally, we qualitatively show how a model trained on StandardSim performs on a real change detection example in Figure \ref{fig:real_example}. The left and middle images correspond to before and after frames, and items are put on the shelf on the bottom right corner. The model's output is highlighted in the rightmost image, and shows that the synthetic to real domain transfer is achievable.
%%%%%%%%%%%%%%%%%%%%%%%%%%%%%%%%%%%%%%%%%%%%%%%
% Table with quantitative change det results
%%%%%%%%%%%%%%%%%%%%%%%%%%%%%%%%%%%%%%%%%%%%%%%
\begin{table}[!t]
\caption{ Class Breakdown (\% IOU)}
\label{tab:change_det_results_breakdown}
\setlength\tabcolsep{4pt}
\centering
\begin{tabular}{|l|l|l|l|l|}
\hline
Class & No change & Put & Take & Shift \\
\hline
Val & 98.30  & 24.46 & 19.30 & 2.56 \\
Test & 96.45 & 25.04 & 19.85 & 2.83 \\
\hline
\end{tabular}
\end{table}
%%%%%%%%%%%%%%%%%%%%%%%%%%%%%%%%%%%%%%%%%%%%%%%
% Figure showing change detection results
%%%%%%%%%%%%%%%%%%%%%%%%%%%%%%%%%%%%%%%%%%%%%%%
\begin{figure}[!t]
\label{fig:results_change_detection}
\caption{
       Qualitative results of our change detection model. The leftmost column shows the initial scene, the second column shows the scene after changes. The third column shows the model's predictions and the fourth column shows the ground truth label. Green signifies puts, red are takes and purple are shifts. Best viewed in color and zoomed in.
}
\tabcolsep 0.03cm
\makebox[\textwidth]{
\begin{tabular}{ccccccc}
  \includegraphics[trim = 0mm 0mm 0mm 0mm,clip,height=1.5cm]{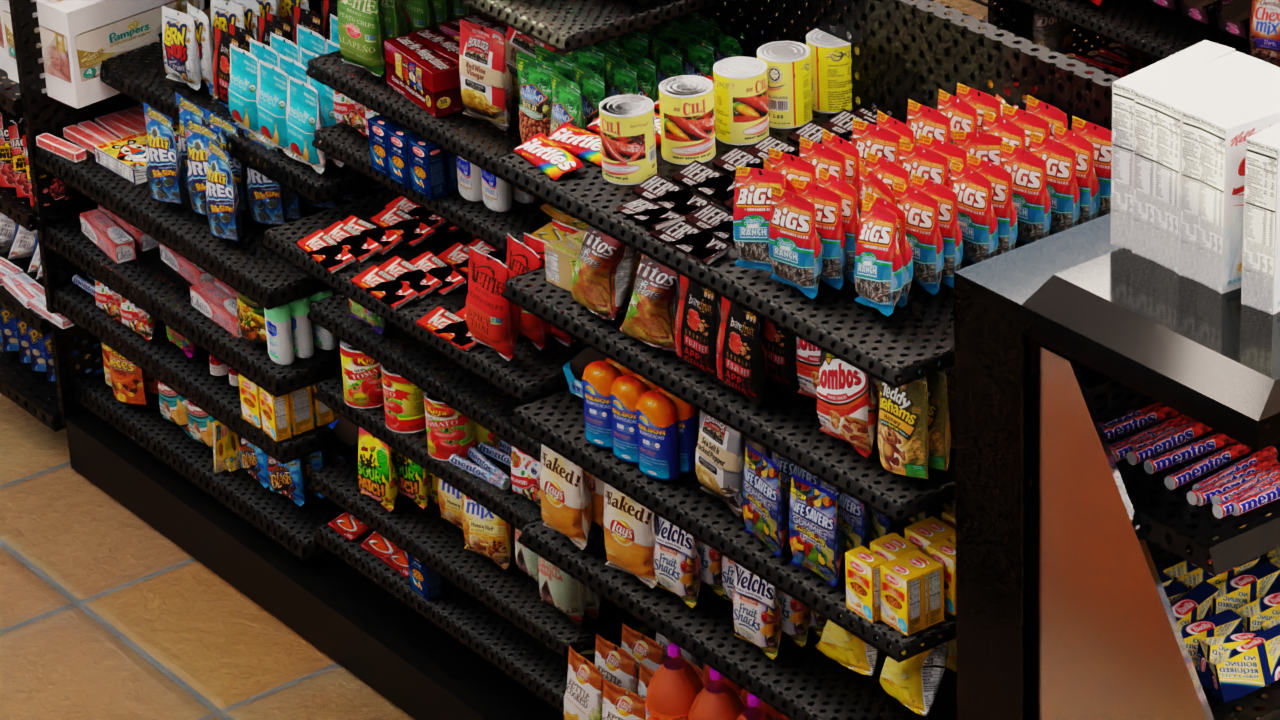} &
  \includegraphics[trim = 0mm 0mm 0mm 0mm,clip,height=1.5cm]{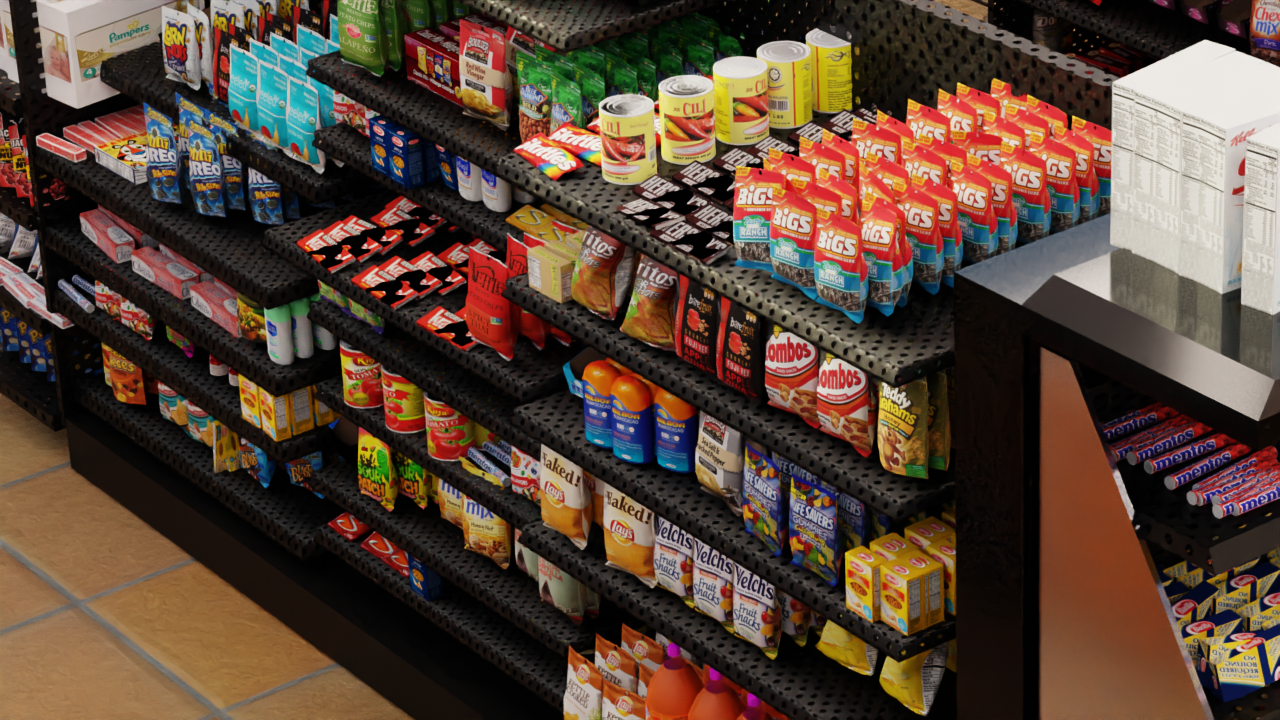} &
  \includegraphics[trim = 0mm 0mm 0mm 0mm,clip,height=1.5cm]{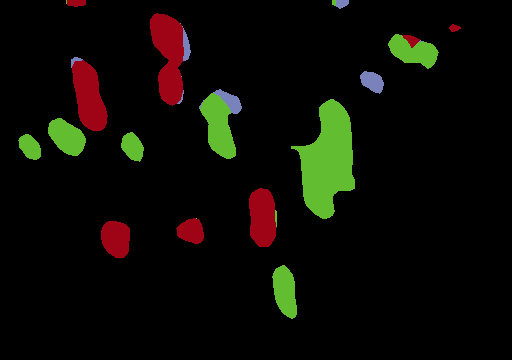} &
  \includegraphics[trim = 0mm 0mm 0mm 0mm,clip,height=1.5cm]{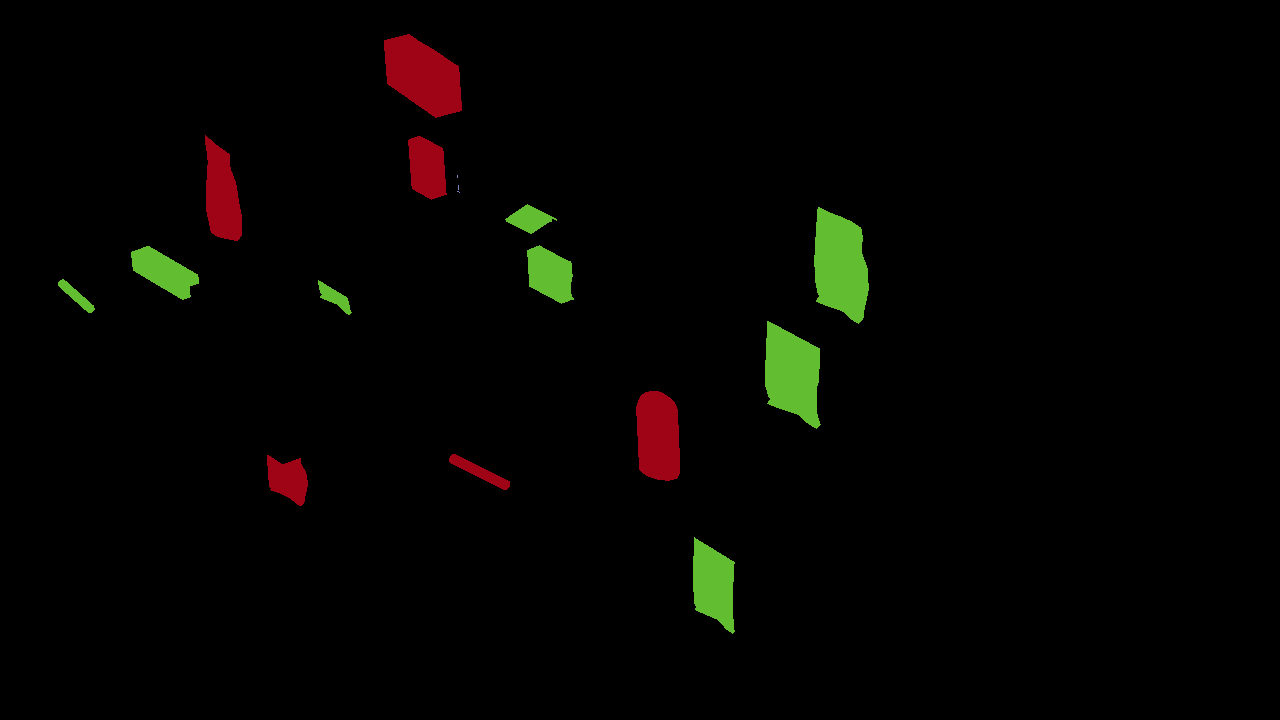} \\
  
  \includegraphics[trim = 0mm 0mm 0mm 0mm, clip, height=1.5cm]{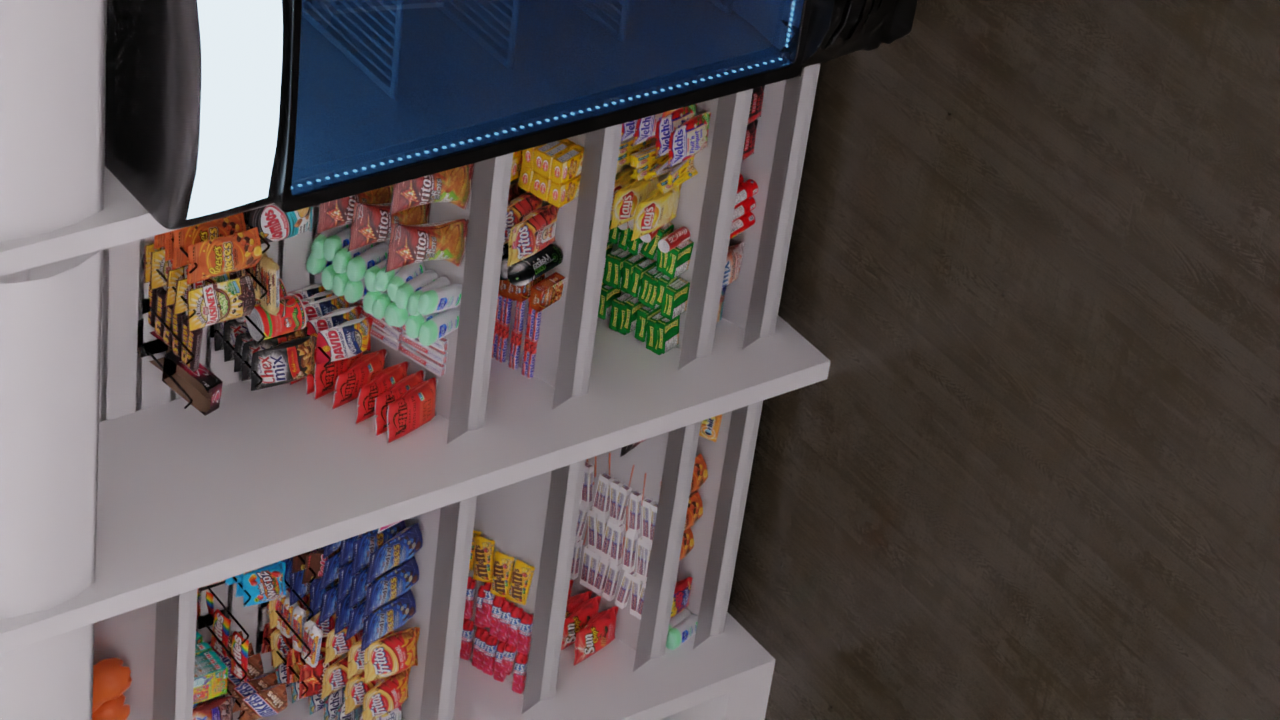} &
  \includegraphics[trim = 0mm 0mm 0mm 0mm, clip, height=1.5cm]{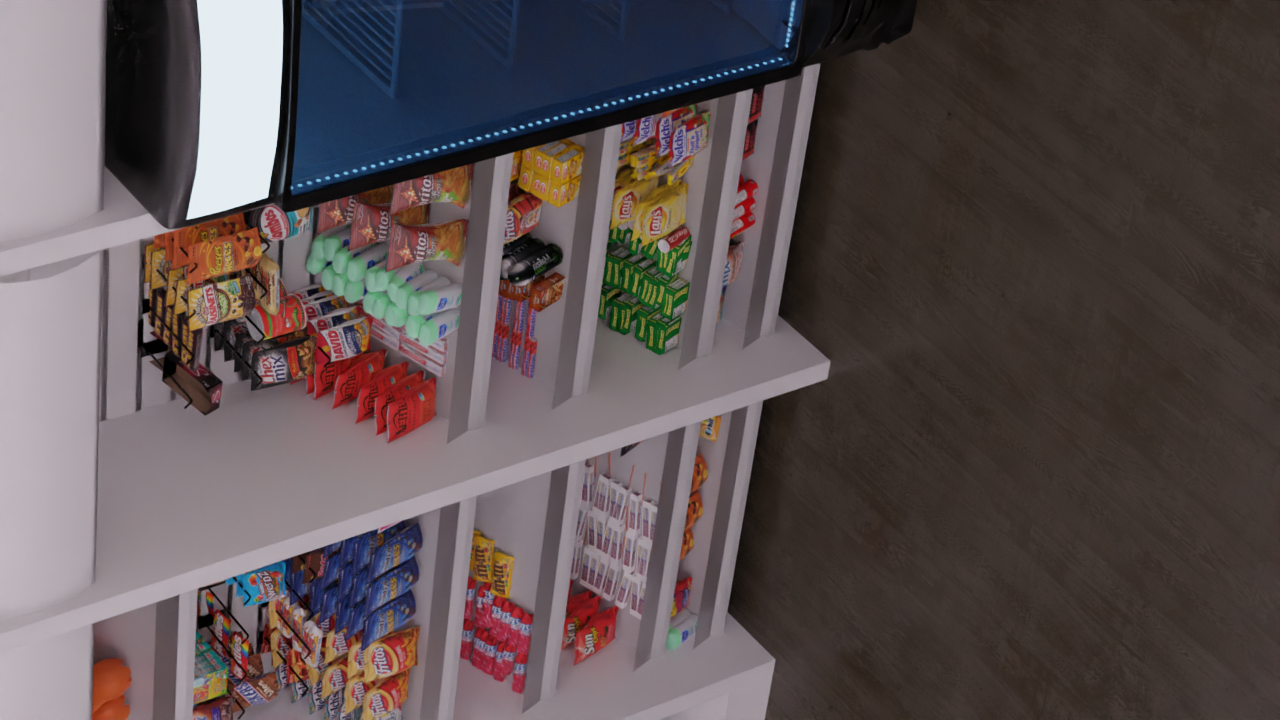} & 
  \includegraphics[trim = 0mm 0mm 0mm 0mm, clip, height=1.5cm]{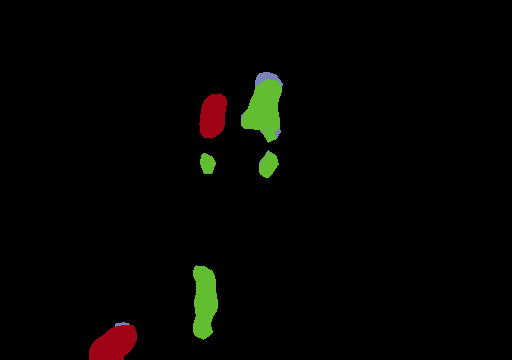} &
\includegraphics[trim = 0mm 0mm 0mm 0mm, clip, height=1.5cm]{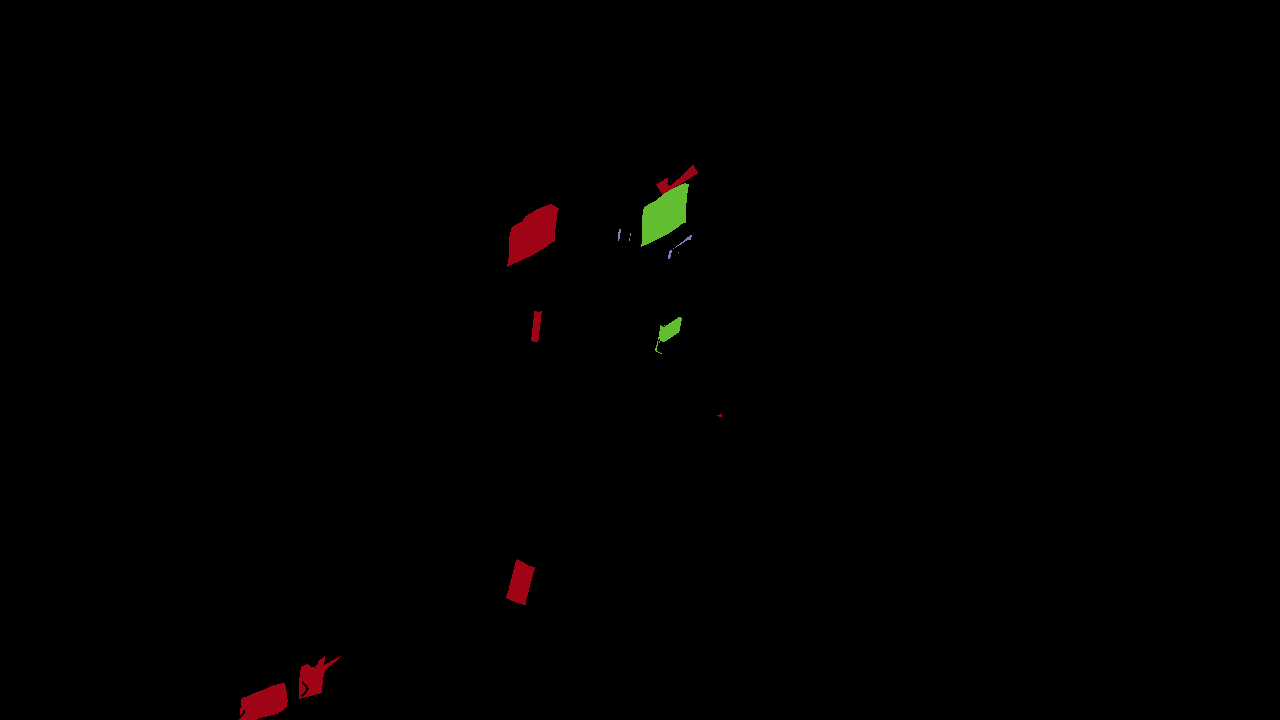} &
\end{tabular}
}
\end{figure}
%%%%%%%%%%%%%%%%%%%%%%%%%%%%%%%%%%%%%%%%%%%%%%%
% Figure showing change det on a real sample
%%%%%%%%%%%%%%%%%%%%%%%%%%%%%%%%%%%%%%%%%%%%%%%
\begin{figure}[!t]
\label{fig:real_example}
\caption{
       Three frames from a real scene of a retail environment where changes occur in the bottom right corner of the scene. The left and middle images show the before and after frames. On the right the output from a model trained on StandardSim is overlaid on top.
}
\tabcolsep 0.03cm
\makebox[\textwidth]{
\begin{tabular}{ccc}
  \includegraphics[trim = 0mm 0mm 0mm 0mm,clip,height=2.2cm]{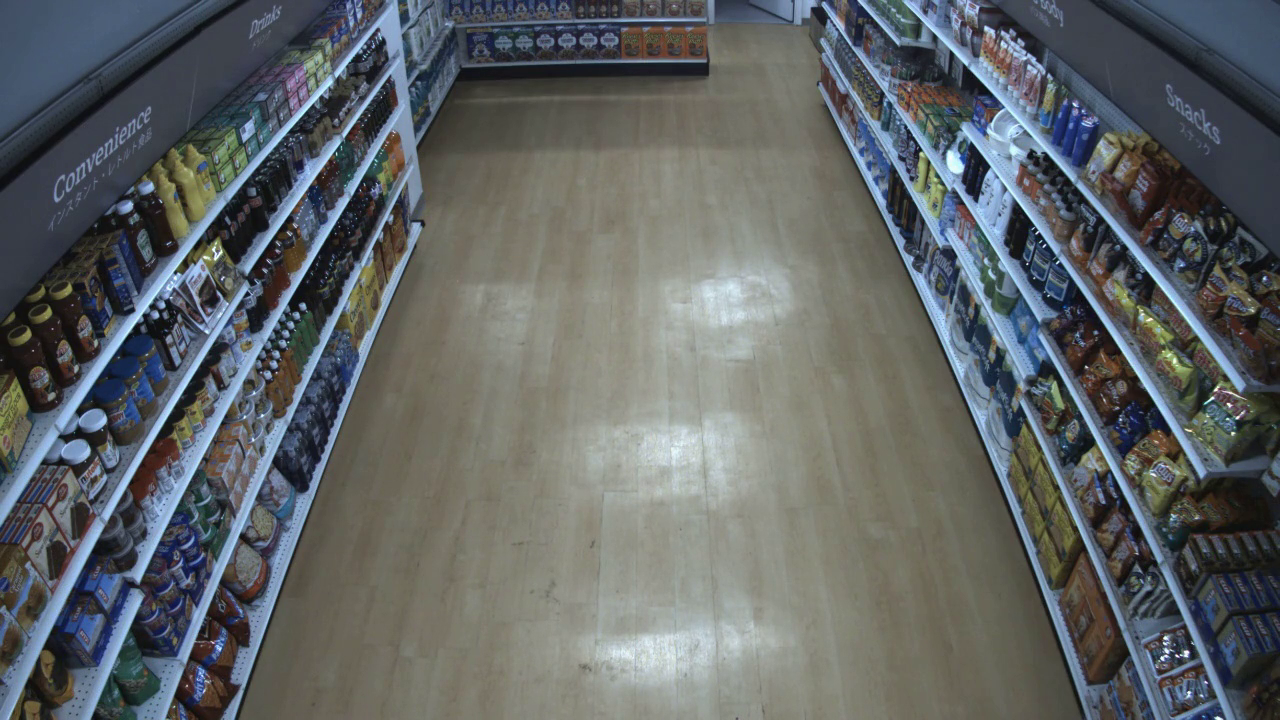} &
  \includegraphics[trim = 0mm 0mm 0mm 0mm, clip, height=2.2cm]{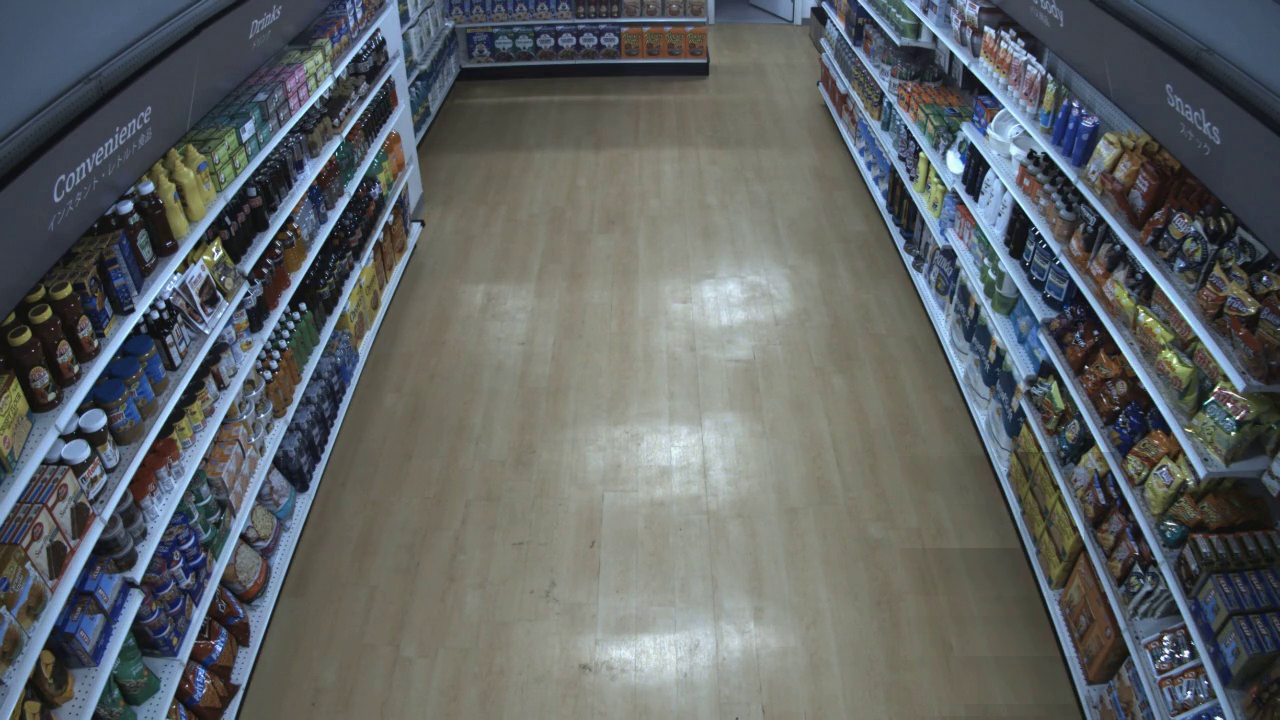} &
  \includegraphics[trim = 0mm 0mm 0mm 0mm,clip,height=2.2cm]{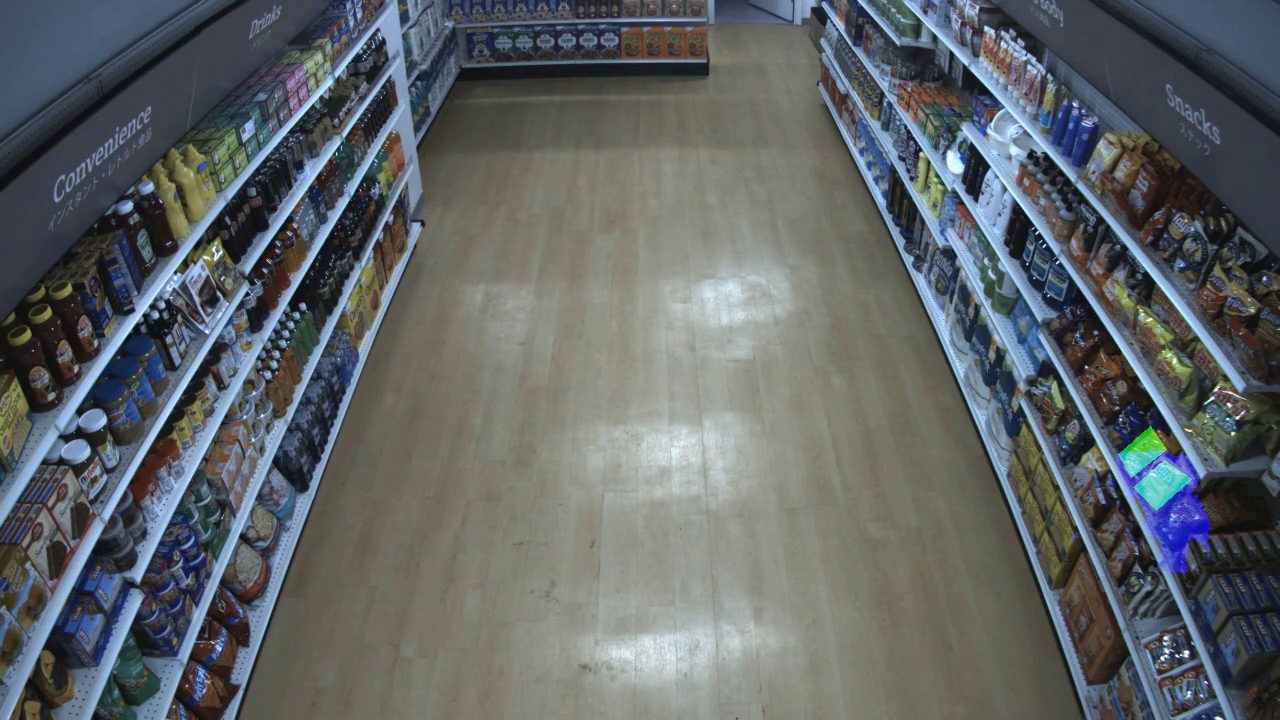} \\
\end{tabular}}
\end{figure}
\subsection{Depth Estimation}
\par Many datasets for monocular depth estimation exist and have been used to train the state of the art monocular depth estimation method MiDaS \cite{Ranftl2020}. The first version of MiDaS took advantage of 3D movies from which it reconstructed depth from neighboring frames. Their code provides an inference script and trained weights for their models, but no training script or implementation of their loss functions. The next iteration of their model builds off of recent developments in transformers for vision, and they develop Dense Prediction Transformers \cite{Ranftl2021} for several dense prediction tasks including monocular depth estimation. We benchmark their DPT-Large model that was pretrained on their MIX 6 meta-dataset and keep their model parameters unchanged. We find that on our test set it achieves an absolute relative error of 83.59, and absolute difference of 10.03 and a square relative error of 625.80.
\par We note that in some of our scenes, there are areas of the camera in that are blocked by a lighting fixture, and this results in a ground truth depth of 0 meters. To prevent NANs in the metric calculations we masked out the pixels where ground truth is 0, which amounted to ignoring 0.089\% of the pixels. This is a negligible amount and unlikely to sway the overall metrics.
\par Table \ref{tab:depth_results_comparison} compares the model's performance on our test set versus other datasets and shows that its performance is considerably worse on our dataset in absolute relative error. We include some qualitative results in Figure \ref{fig:results_depth_estimation} on our dataset and on the NYUv2 dataset \cite{SilbermanECCV12}, consisting of indoor samples, as well as on an outdoor sample. The model performs well on NYUv2 and on our dataset except that it cannot distinguish small changes in depth between small objects. This is also apparent from the depth estimation of the tree branches in the rightmost column. Further, when we reconstruct the depth estimate from a 3D perspective in Figure \ref{fig:depth_reconstruction} on our dataset, we find that corners and areas between shelves are inaccurate. This suggests that our dataset may serve as a useful augmentation to their MIX 6 to improve estimation of small and distant objects, especially since our ground truth is accurate.
%%%%%%%%%%%%%%%%%%%%%%%%%%%%%%%%%%%%%%%%%%%%%%%
% Table with quantitative MiDaS performance
%%%%%%%%%%%%%%%%%%%%%%%%%%%%%%%%%%%%%%%%%%%%%%%
\begin{table}[!t]
\caption{Absolute Relative Error (ARE) of DPT-Large on multiple datasets}
\label{tab:depth_results_comparison}
\setlength\tabcolsep{4pt}
\centering
\begin{tabular}{|l|l|l|l|l|l|}
\hline
& Sintel \cite{Butler:ECCV:2012} & TUM \cite{sturm12iros} & Kitti \cite{Uhrig2017THREEDV} & NYUv2 \cite{SilbermanECCV12} & Ours (test) \\
\hline
ARE & 0.27 & 9.97  & 8.46 & 8.43 & 83.59 \\
\hline
\end{tabular}
\end{table}
%%%%%%%%%%%%%%%%%%%%%%%%%%%%%%%%%%%%%%%%%%%%%%%%
% Figure showing qualitative MiDaS depth results
%%%%%%%%%%%%%%%%%%%%%%%%%%%%%%%%%%%%%%%%%%%%%%%%
\begin{figure}[!t]
\caption{
       Comparing DPT-Large results on different datasets. The top row shows the original image and the bottom row shows the output. Columns A and B show a change image pair. The model is able to identify general outlines of boxes well but fails to distinguish small changes in depth between small objects. Column C shows an indoor scene from the NYUv2 dataset and column D shows an outdoor scene. The model struggles with reflective surfaces and fine structures.
}
\label{fig:results_depth_estimation}
\tabcolsep 0.03cm
\noindent\makebox[\textwidth]{
\begin{tabular}{cccc}
  \includegraphics[trim = 0mm 0mm 0mm 0mm,clip,width=3.0cm,height=2.2cm]{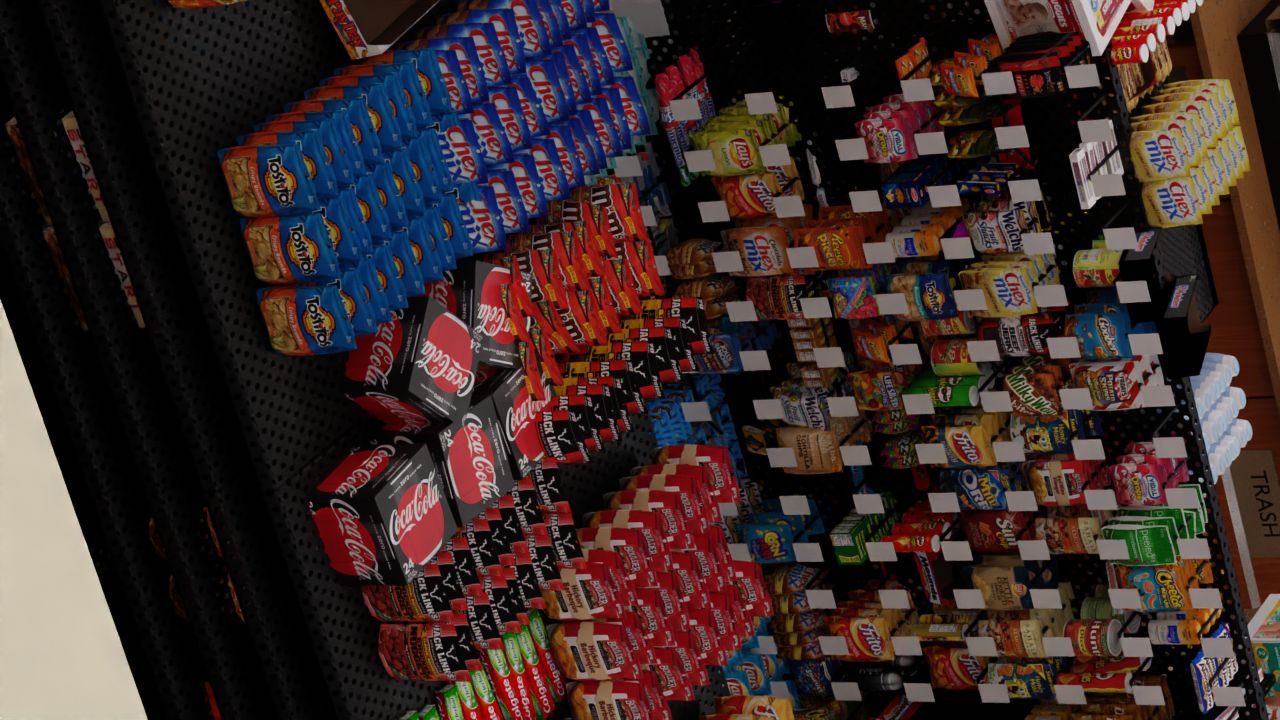} &
  \includegraphics[trim = 0mm 0mm 0mm 0mm, clip, width=3.0cm,height=2.2cm]{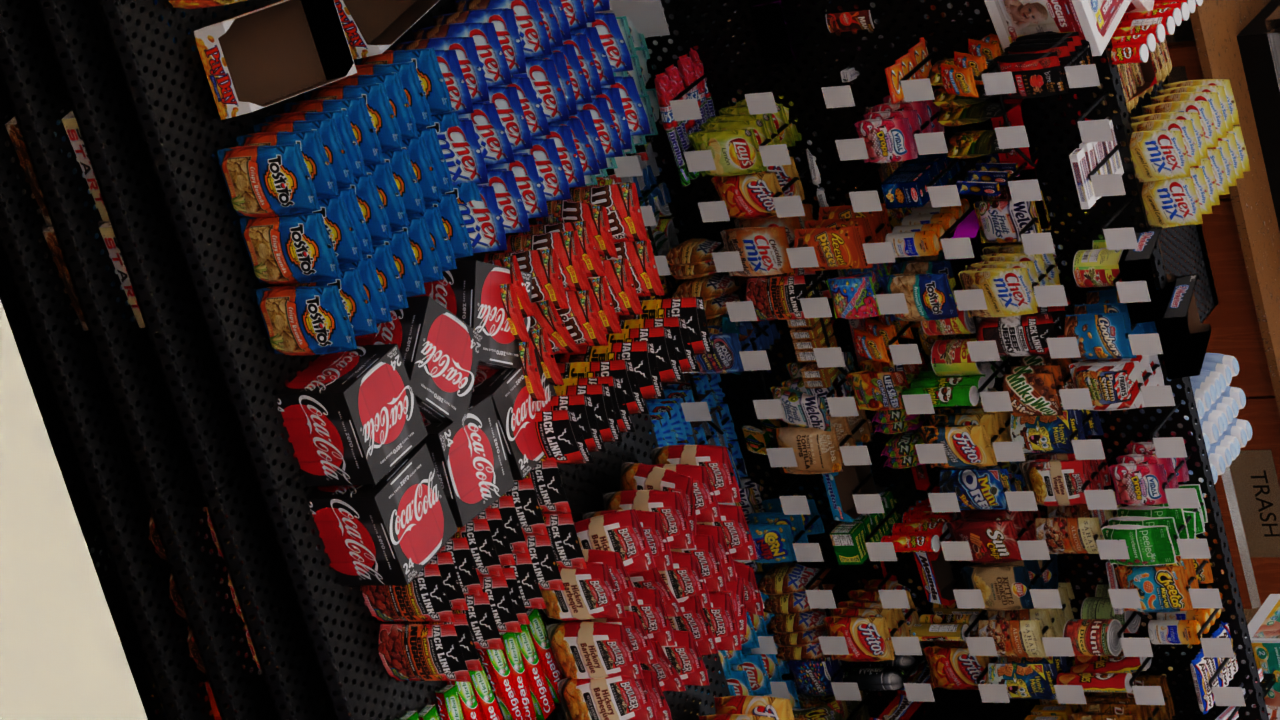} &
  \includegraphics[trim = 0mm 0mm 0mm 0mm, clip, width=3.0cm]{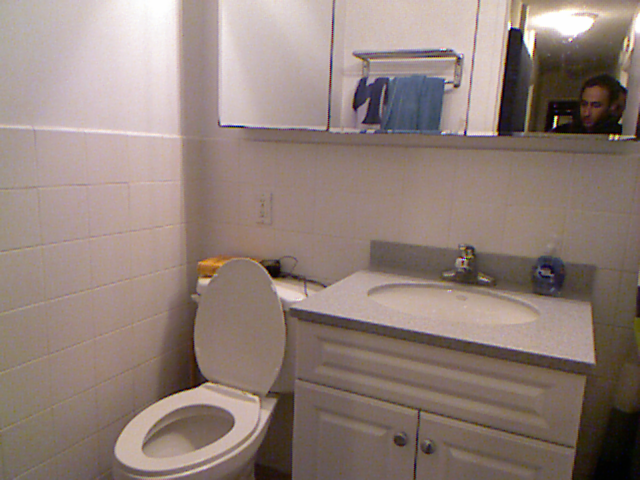} &
  \includegraphics[trim = 0mm 0mm 0mm 0mm, clip, width=3.0cm]{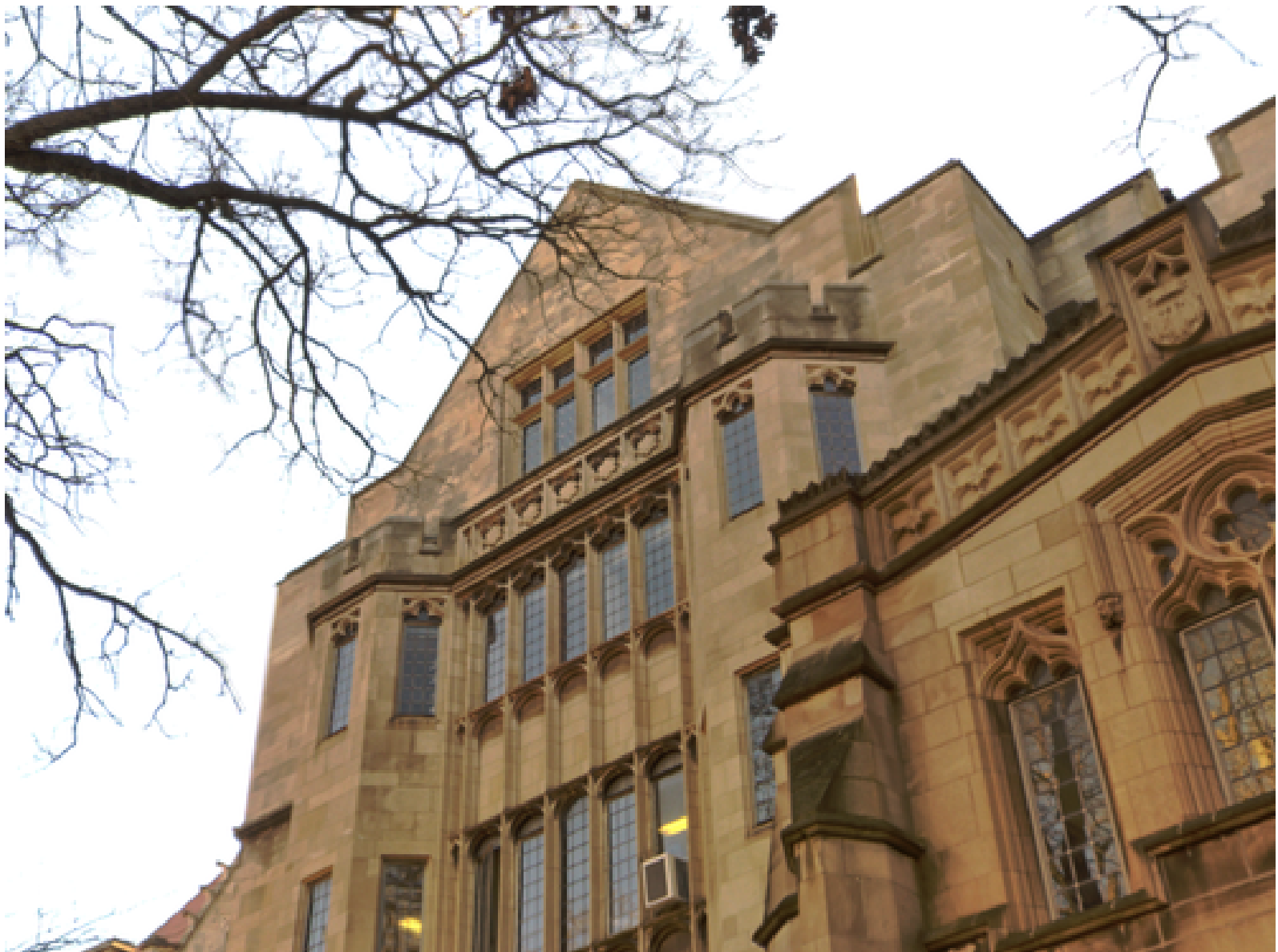} \\
  
  \includegraphics[trim = 0mm 0mm 0mm 0mm,clip,width=3.0cm,height=2.2cm]{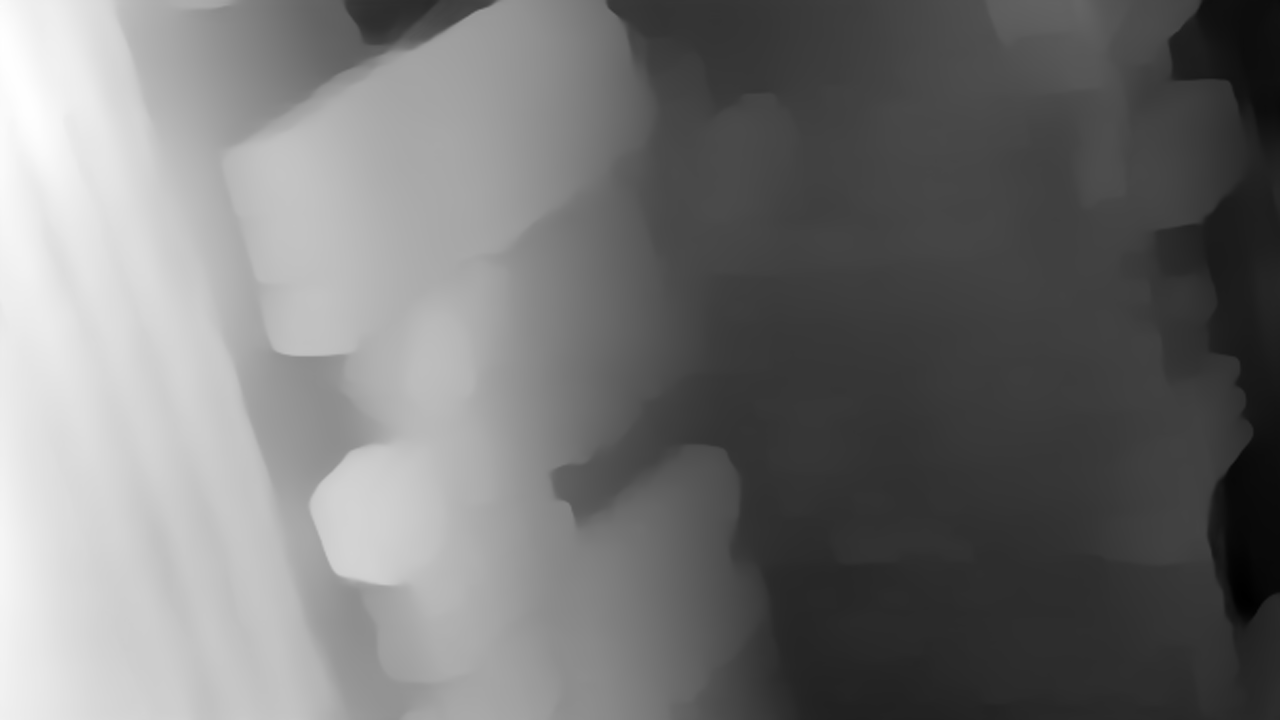} &
  \includegraphics[trim = 0mm 0mm 0mm 0mm, clip, width=3.0cm,height=2.2cm]{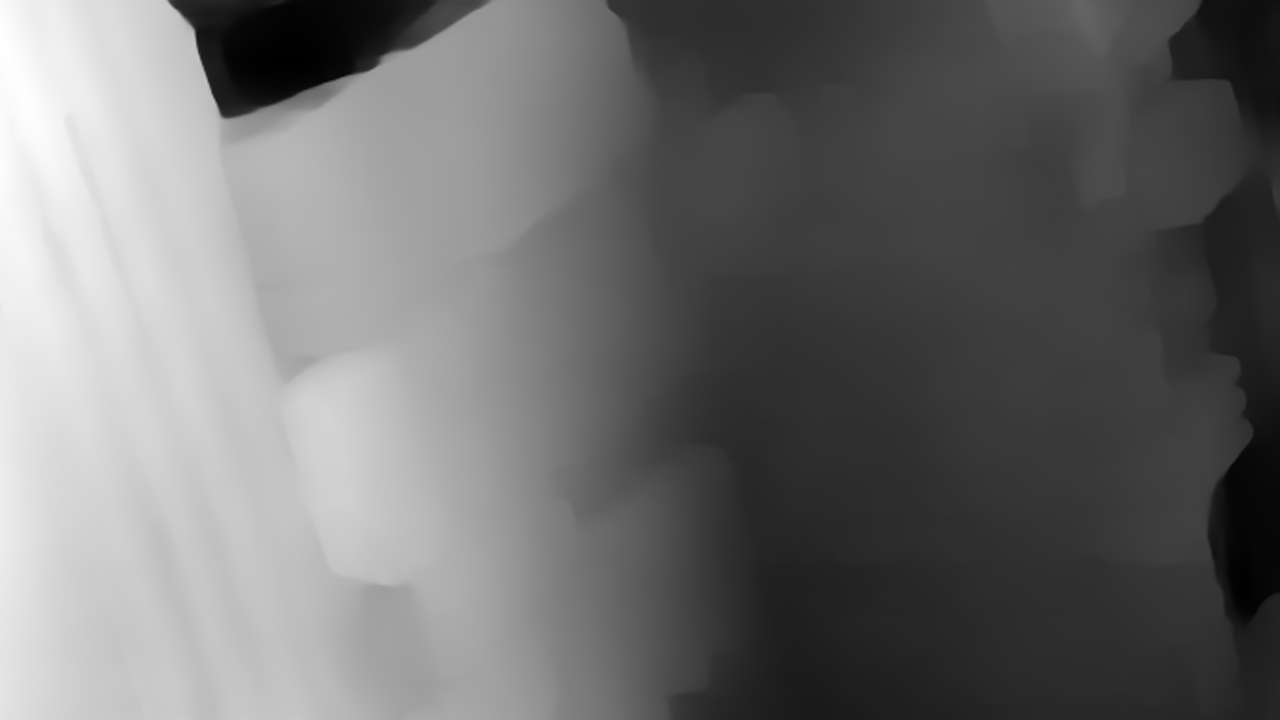} &
  \includegraphics[trim = 0mm 0mm 0mm 0mm, clip, width=3.0cm]{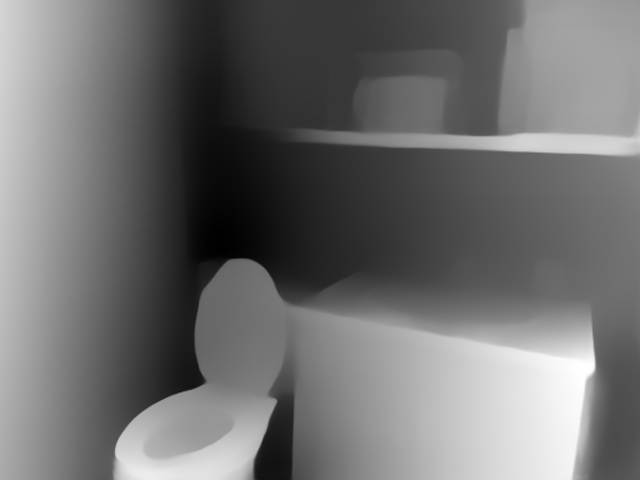} &
  \includegraphics[trim = 0mm 0mm 0mm 0mm, clip, width=3.0cm]{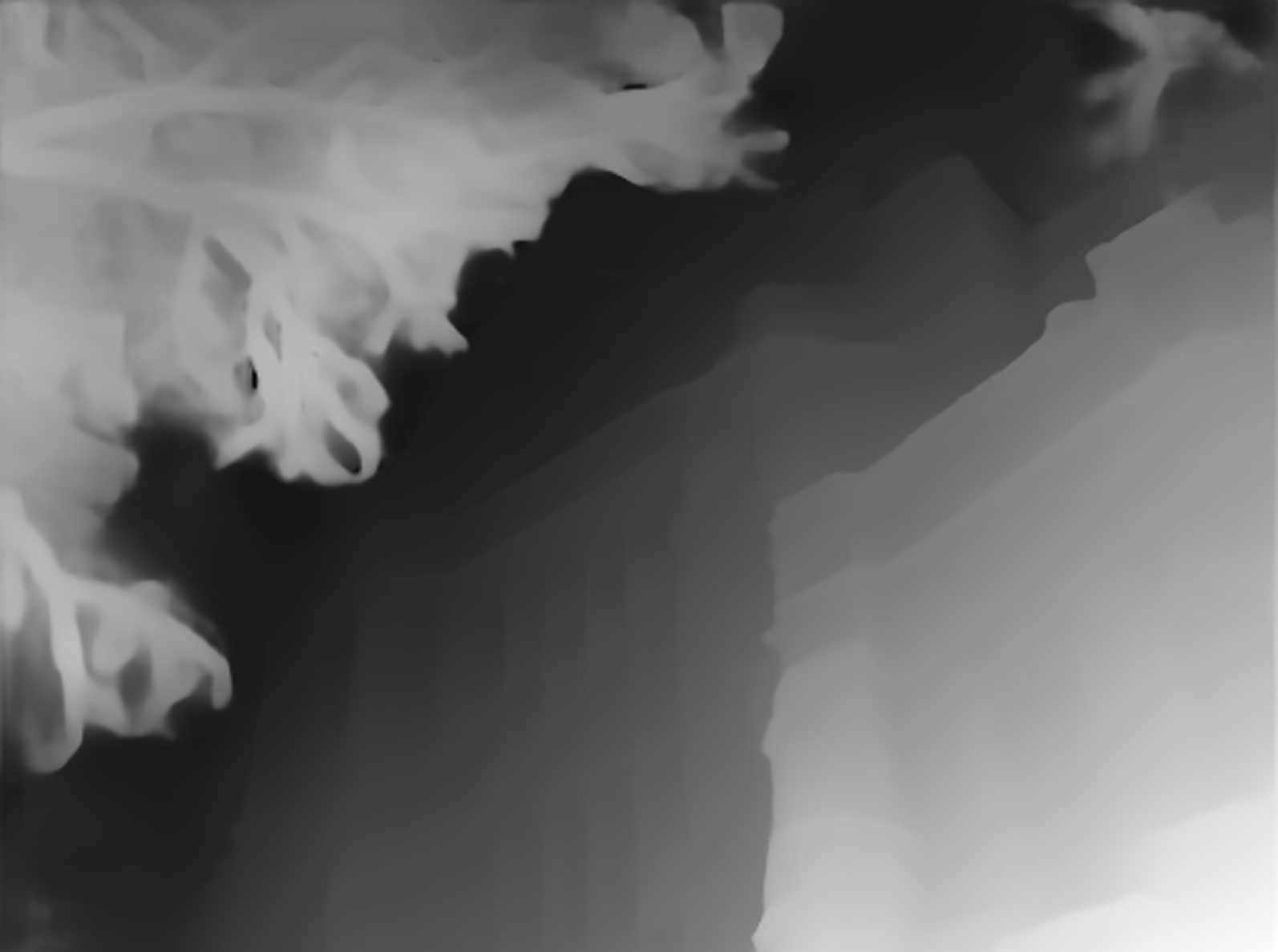} \\

A & B & C & D\\
\end{tabular}}
\end{figure}
%Possible experiment: Training a monocular depth estimation model on KITTI and our dataset and seeing how it performs on KITTI tells us whether our dataset allows for transfer learning or is different enough to constitute a different domain.
%%%%%%%%%%%%%%%%%%%%%%%%%%%%%%%%%%%%%%%%%%%%%%%%
% Figure showing depth reconstruction
%%%%%%%%%%%%%%%%%%%%%%%%%%%%%%%%%%%%%%%%%%%%%%%%
\begin{figure}[!t]
\caption{
       Two depth estimation reconstruction examples. The leftmost column shows the ground truth depth while the second column from the left shows the DPT-Large's prediction reconstruction. The rightmost two images show the ground truth and prediction from a different perspective. The spaces between shelves indicate that corners are challenging for depth estimation.
}
\label{fig:depth_reconstruction}
\tabcolsep 0.03cm
\noindent\makebox[\textwidth]{
\begin{tabular}{c c c c}
  \includegraphics[trim = 0mm 0mm 0mm 0mm,clip,height=2.0cm]{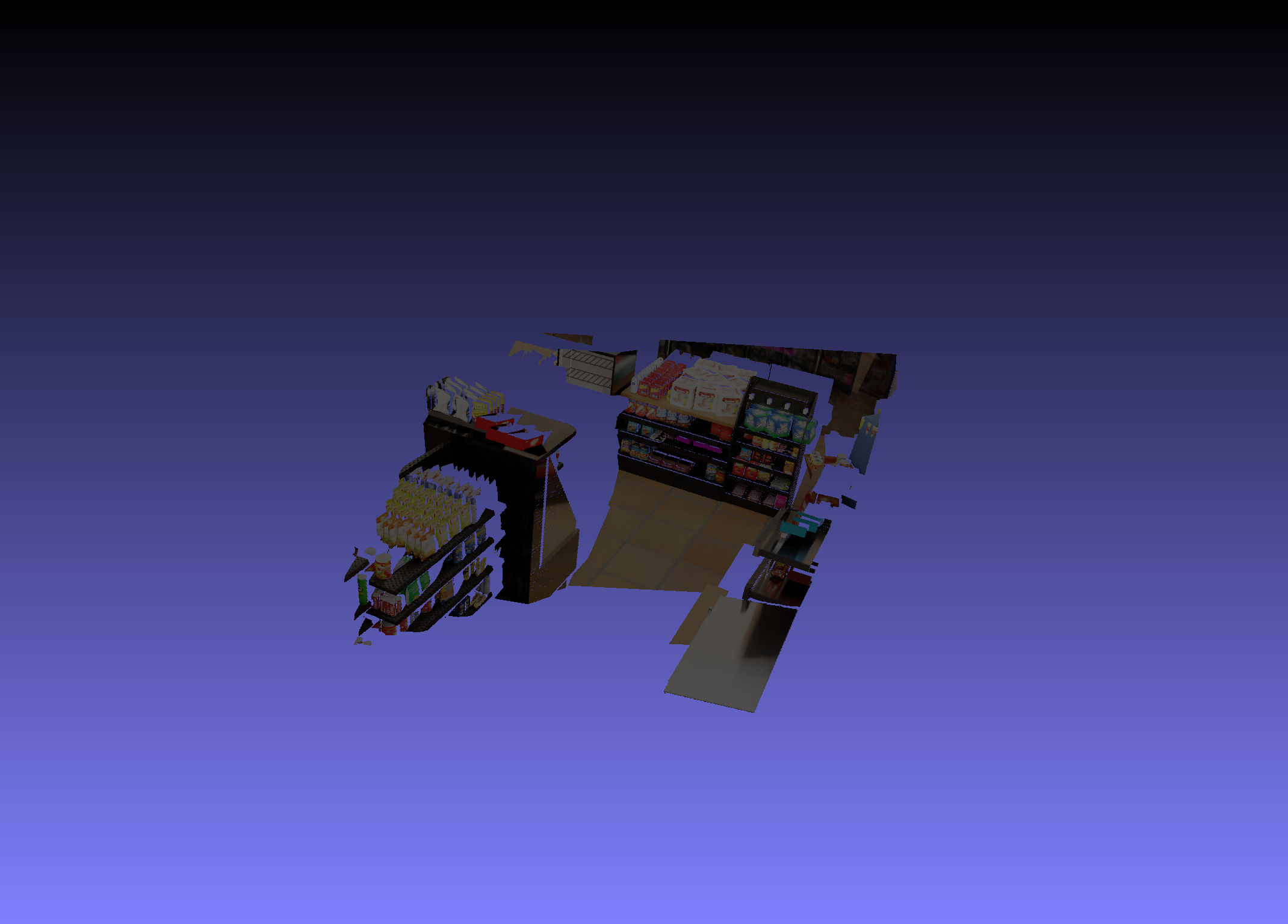} &
  \includegraphics[trim = 0mm 0mm 0mm 0mm, clip, height=2.0cm]{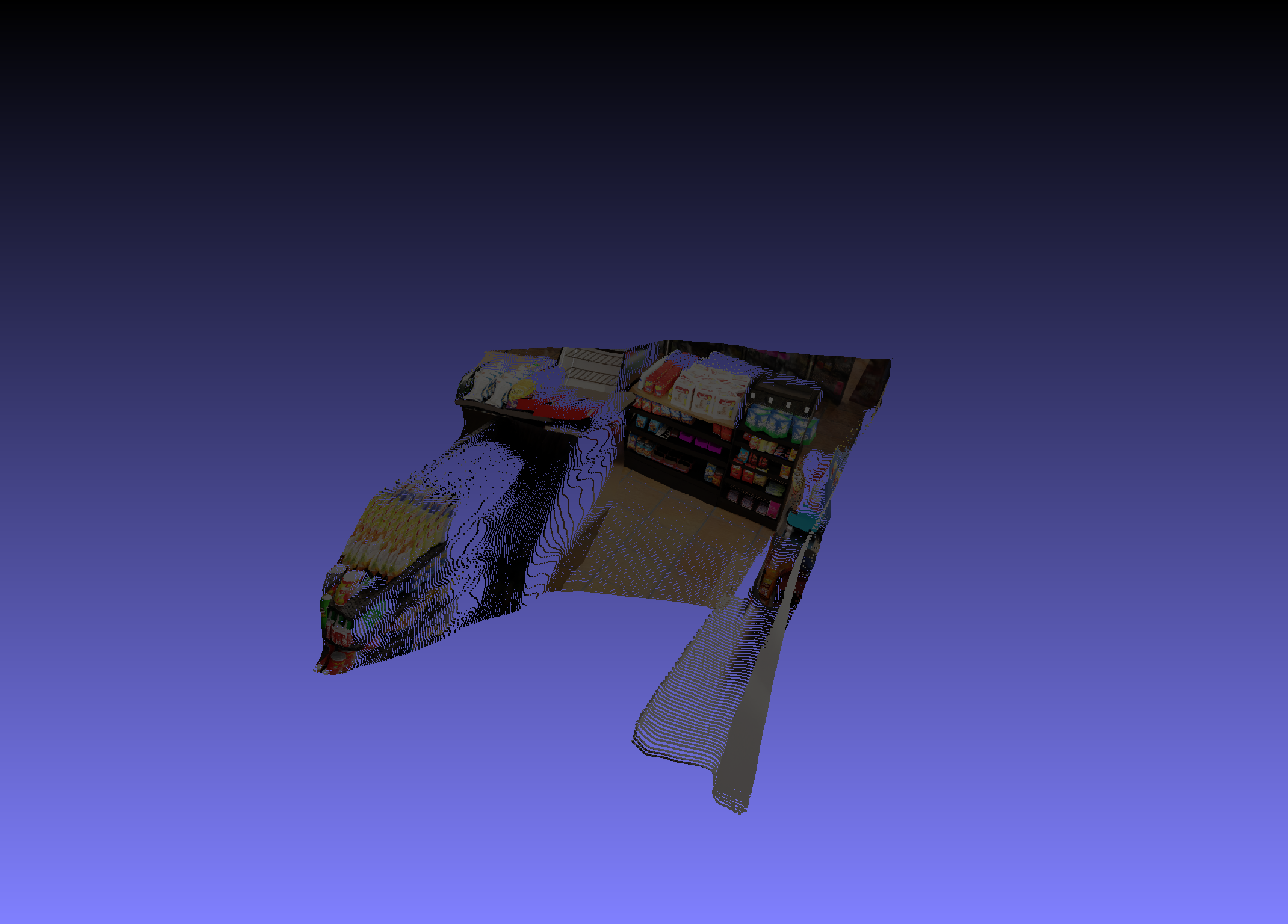} &
  \includegraphics[trim = 0mm 0mm 0mm 0mm,clip,height=2.0cm]{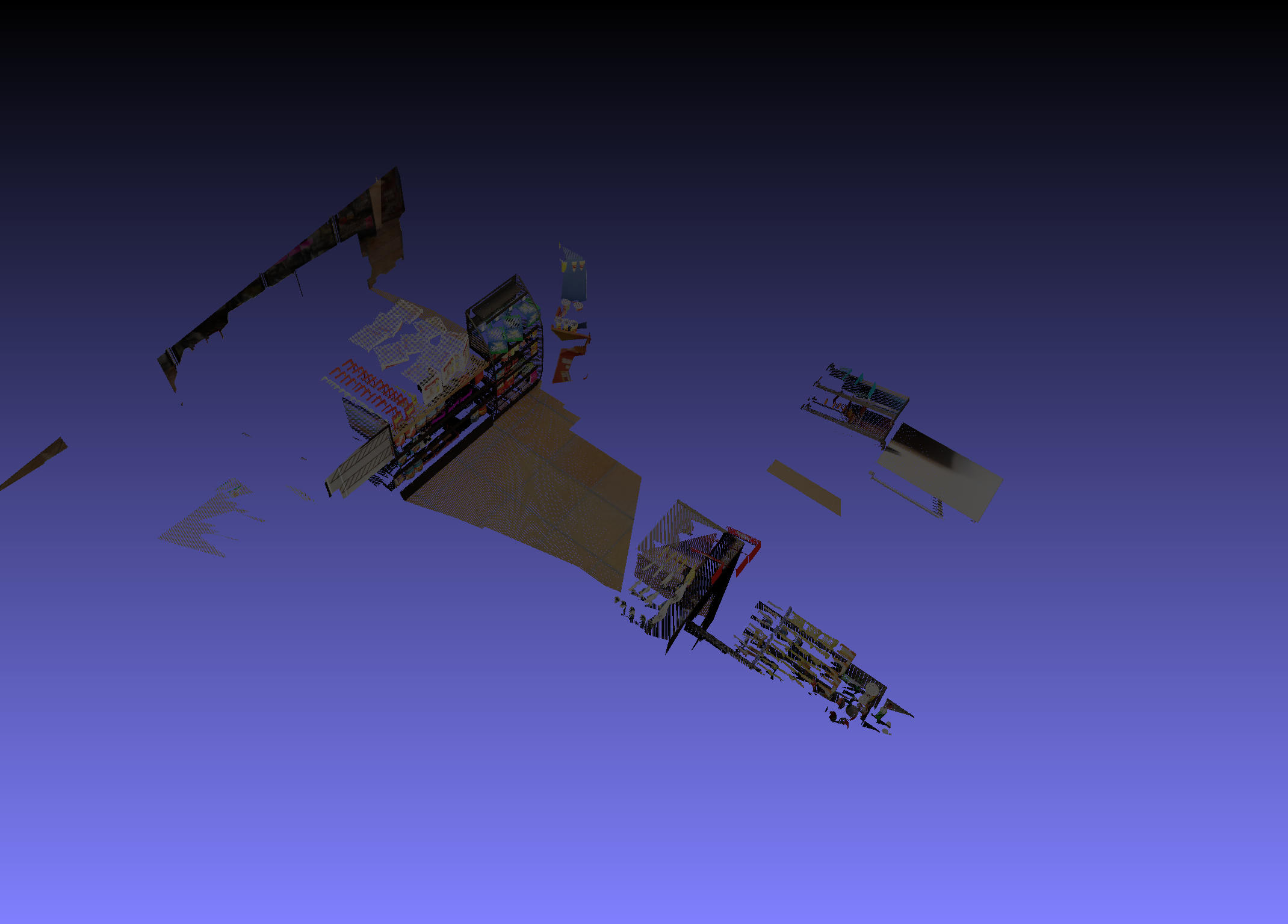} &
  \includegraphics[trim = 0mm 0mm 0mm 0mm, clip, height=2.0cm]{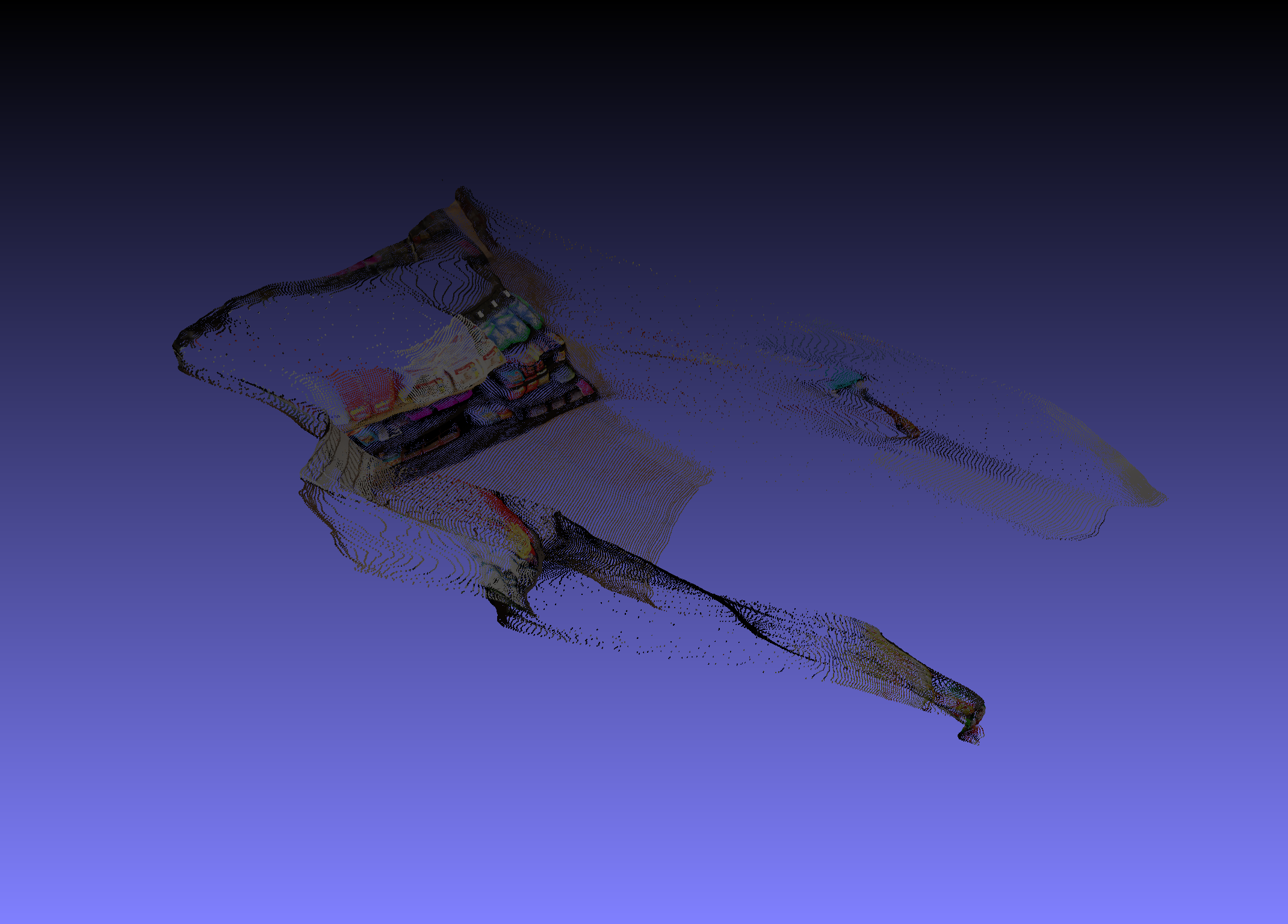} \\
\end{tabular}}
\end{figure}
%%%%%%%%%%%%%%%%%%%%%%%%%%%%%%%%%%%%%%%%%%%%%%%%
% Figure showing multiple cam views
%%%%%%%%%%%%%%%%%%%%%%%%%%%%%%%%%%%%%%%%%%%%%%%%
\begin{figure}[!t]
\caption{
       Our dataset offers multiple views of each scene. Here we have the same scene rendered from three different cameras.
}
\label{fig:multi_cam_example}
\tabcolsep 0.03cm
\noindent\makebox[\textwidth]{%
\begin{tabular}{ccc}
\includegraphics[trim = 0mm 0mm 0mm 0mm,clip,height=2.0cm]{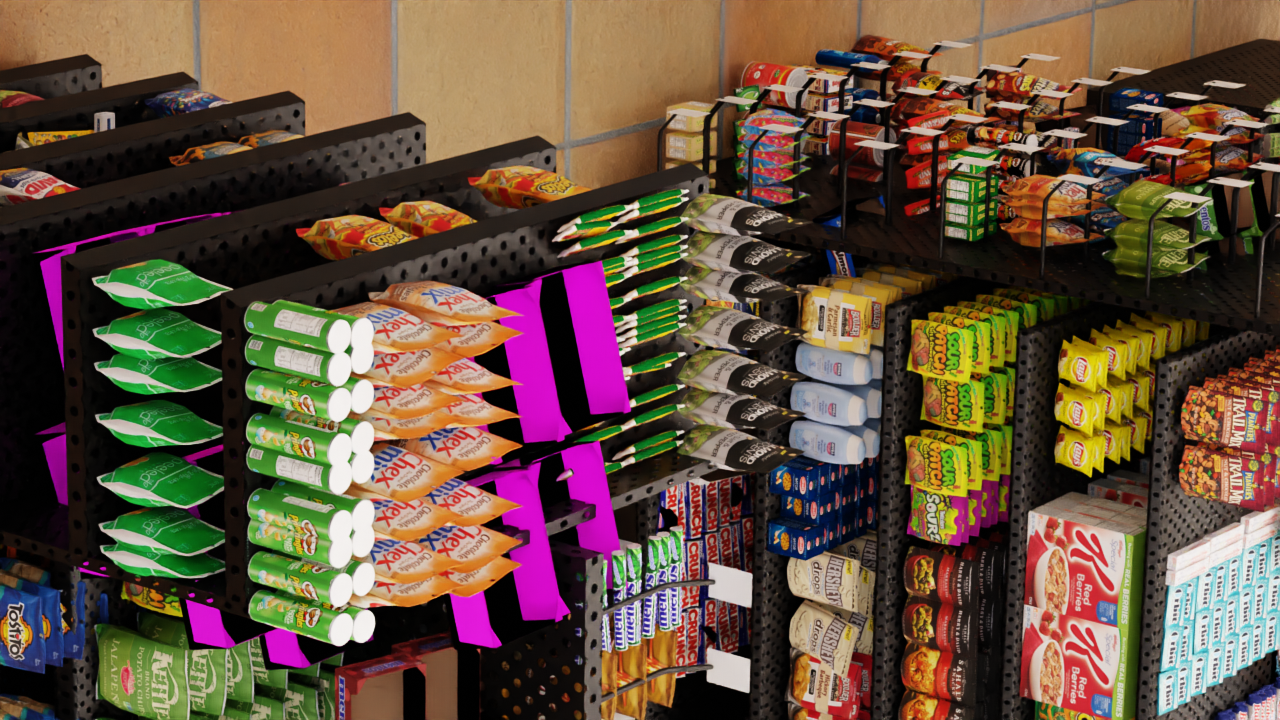} &
\includegraphics[trim = 0mm 0mm 0mm 0mm,clip,height=2.0cm]{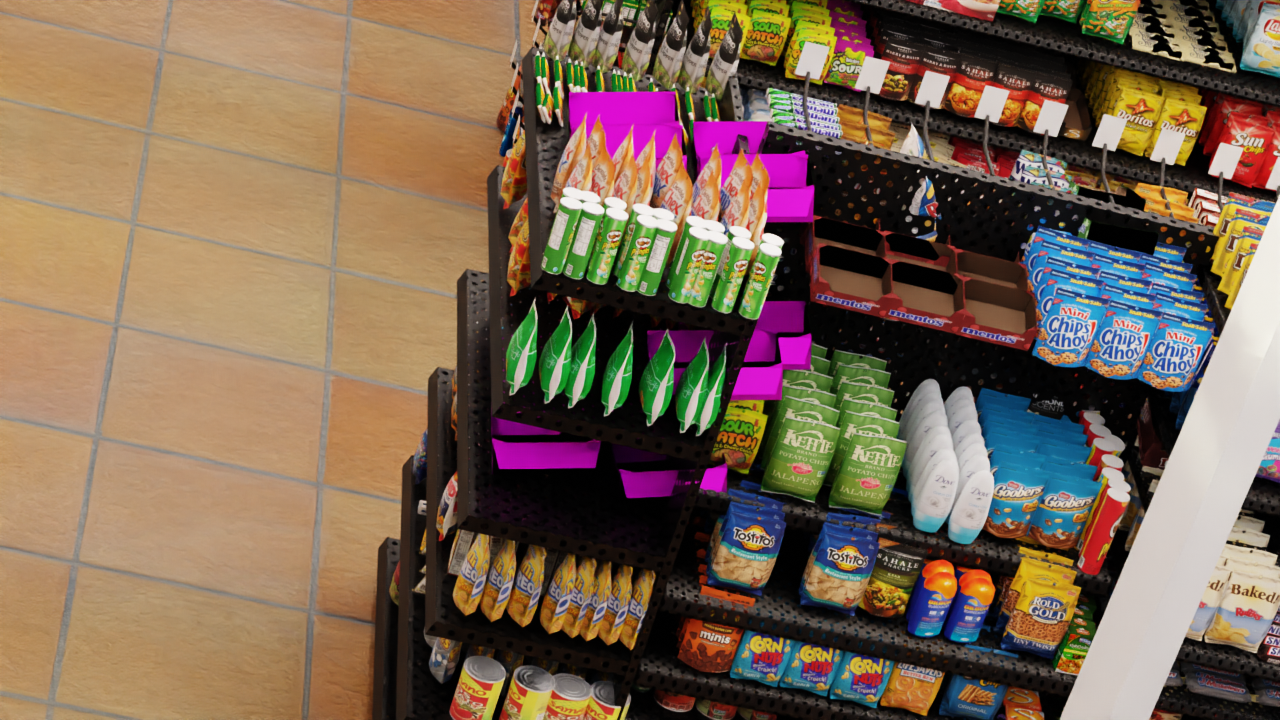} &
\includegraphics[trim = 0mm 0mm 0mm 0mm, clip, height=2.0cm]{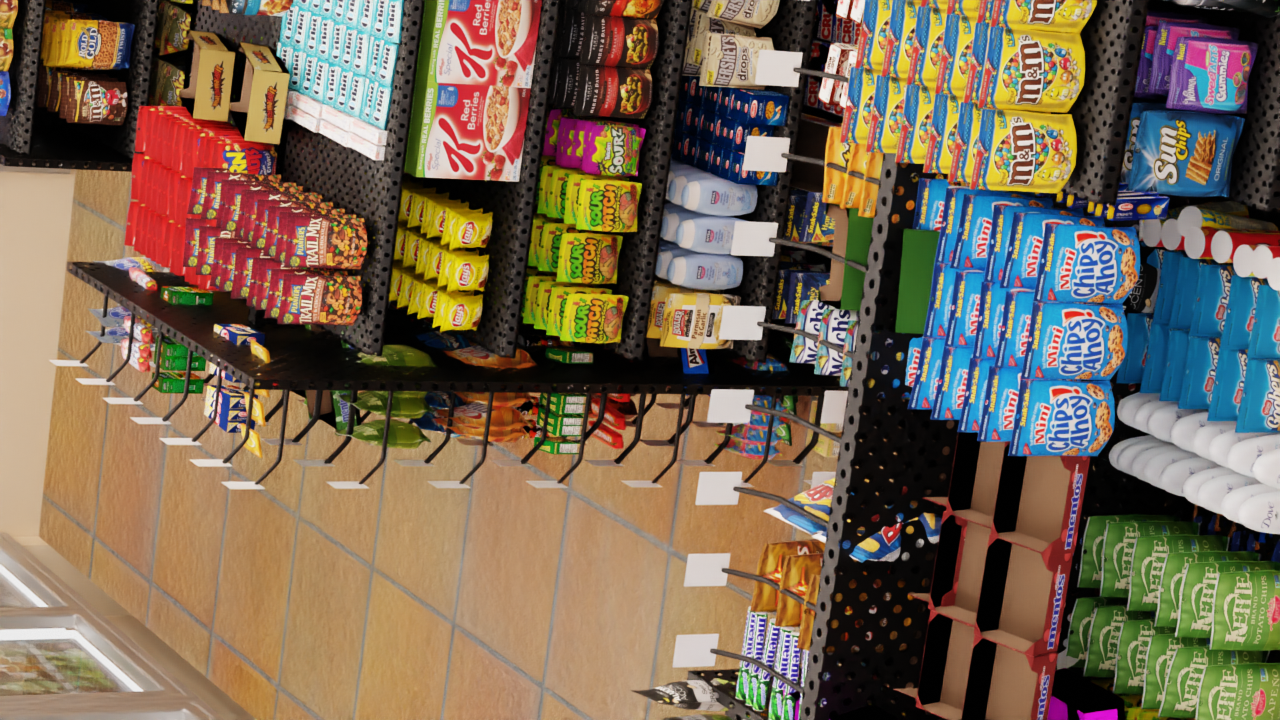} \\

\end{tabular}}
\end{figure}
%%%%%%%%%%%%%%%%%%%%%%%%%%%%%%%%%%%%%%%%%%%%%%%%
% Discussion
%%%%%%%%%%%%%%%%%%%%%%%%%%%%%%%%%%%%%%%%%%%%%%%%
\section{Discussion}
%Discuss decision for handling pixels in change detection that count as more than one class.
%Discussion around how MIDAS performs poorly on our dataset 
Top-Down cameras present difficulties for models trained on existing datasets. We parameterize the angle of tilt toward the ground as alpha, where 0 means pointed at the horizon, 90 means pointed straight down, and -90 means pointed straight up. Self-driving datasets are gathered from dashboard videos like Kitti have an alpha of 0. Datasets gathered from movies and disparity maps have limited alpha range as well. Our dataset has an alpha range of [30, 90].
\par Additionally, many datasets are gathered in outdoor settings, which means that the pixels near the top are the sky and have a filler infinite value. Our dataset generally has ceilings in the upper part of the image rather than a sky which can present difficulties for models trained on outside datasets.
\par One specific task that we have not explored in our dataset, but which is possible with the data provided, is that of multiview 3D reconstruction \cite{Hartley2004}. Since each scene has 3 different views along with associated intrinsic and extrinsic calibrations for the cameras, it is possible to use either classical computer vision methods \cite{Hartley2004} or deep learning methods to reconstruct the full 3D scene \cite{mvdepthnet}. This has potential applications to many of the problems discussed above, allowing for 3D object detection and instance segmentation \cite{chen2017multi} of items if the views are detailed enough. 3D reconstructions can also provide information about store layout changes.
\par Labels for instance segmentation can be generated using the polygons provided from in the annotation files. The relevant functions are provided in the repository, along with some simple examples to set up for the instance segmentation task. Detailed item labels can be generated, as well as larger item group labels, such as labels for a group of candy bars.
%%%%%%%%%%%%%%%%%%%%%%%%%%%%%%%%%%%%%%%%%%%%%%%%
% Conclusion
%%%%%%%%%%%%%%%%%%%%%%%%%%%%%%%%%%%%%%%%%%%%%%%%
\section{Conclusion}
\par In this paper we analyze why progress in autonomous checkout systems has been slow even though major advances continue to be made in computer vision. We conclude that this lag in progress is due to a lack of datasets for retail, and address this by presenting StandardSim, a large-scale synthetic open dataset with annotations for a variety of computer vision tasks. We also introduce the change detection task for retail, where pixel-wise labels that identify semantic changes in the environment for image pairs are tailored to shopper interactions with shelves. Finally, we benchmark state of the art models for change detection and monocular depth estimation on our dataset and compare their performance to that on other datasets. We conclude that StandardSim's retail environment domain is unique and challenging compared to other datasets, and identify further applications where it can be used.
\bibliographystyle{splncs04}
\bibliography{standardsimbib}

\end{document}